\documentclass[final]{cvpr}

\usepackage{times}
\usepackage{epsfig}
\usepackage{graphicx}
\usepackage{amsmath}
\usepackage{amssymb}

\usepackage{comment}
\usepackage{color}
\usepackage{booktabs}
\usepackage{bm}
\usepackage{color}
\usepackage{algorithm,algorithmic} 
\usepackage{epsfig}
\usepackage{caption} 
\usepackage{subfigure}
\usepackage{lipsum}
\usepackage{epstopdf}
\usepackage{array} 
\usepackage{bbding}
\usepackage[american]{babel} 
\usepackage{cite}
\usepackage{CJK}
\usepackage{footnote}
\usepackage{stfloats}
\usepackage{float}
\usepackage{mathrsfs}
\usepackage[section]{placeins}

\usepackage[pagebackref=true,breaklinks=true,colorlinks,bookmarks=false]{hyperref}

\def\cvprPaperID{7596} 
\def\confYear{CVPR 2021}
\begin{document}
	
	\title{
		ID-Unet: Iterative Soft and Hard Deformation for View Synthesis}
	\author{{Mingyu Yin$^{1}$ \quad Li Sun$^{1, 2}$\thanks{Corresponding author, email: sunli@ee.ecnu.edu.cn.
Supported by the the Science and Technology Commission of Shanghai Municipality (No.19511120800).} \quad Qingli Li$^{1}$}\\
    {$^{1}$ Shanghai Key Laboratory of Multidimensional Information Processing, } \\ {$^{2}$Key Laboratory of Advanced Theory and Application in Statistics \& Data Science,} \\ {East China Normal University, 200241 Shanghai, China} \\
}

	\maketitle

	\begin{abstract}
		View synthesis is usually done by an autoencoder, in which the encoder maps a source view image into a latent content code, and the decoder transforms it into a target view image according to the condition. However, the source contents are often not well kept in this setting, which leads to unnecessary changes during the view translation. Although adding skipped connections, like Unet, alleviates the problem, but it often causes the failure on the view conformity. This paper proposes a new architecture by performing the source-to-target deformation in an iterative way. Instead of simply incorporating the features from multiple layers of the encoder, we design soft and hard deformation modules, which warp the encoder features to the target view at different resolutions, and give results to the decoder to complement the details. Particularly, the current warping flow is not only used to align the feature of the same resolution, but also as an approximation to coarsely deform the high resolution feature. Then the residual flow is estimated and applied in the high resolution, so that the deformation is built up in the coarse-to-fine fashion. To better constrain the model, we synthesize a rough target view image based on the intermediate flows and their warped features. The extensive ablation studies and the final results on two different data sets show the effectiveness of the proposed model. \url{https://github.com/MingyuY/Iterative-view-synthesis}
	\end{abstract}
	
	\section{Introduction}
	\begin{figure}
		\begin{center} 
			\includegraphics[height=3.9cm]{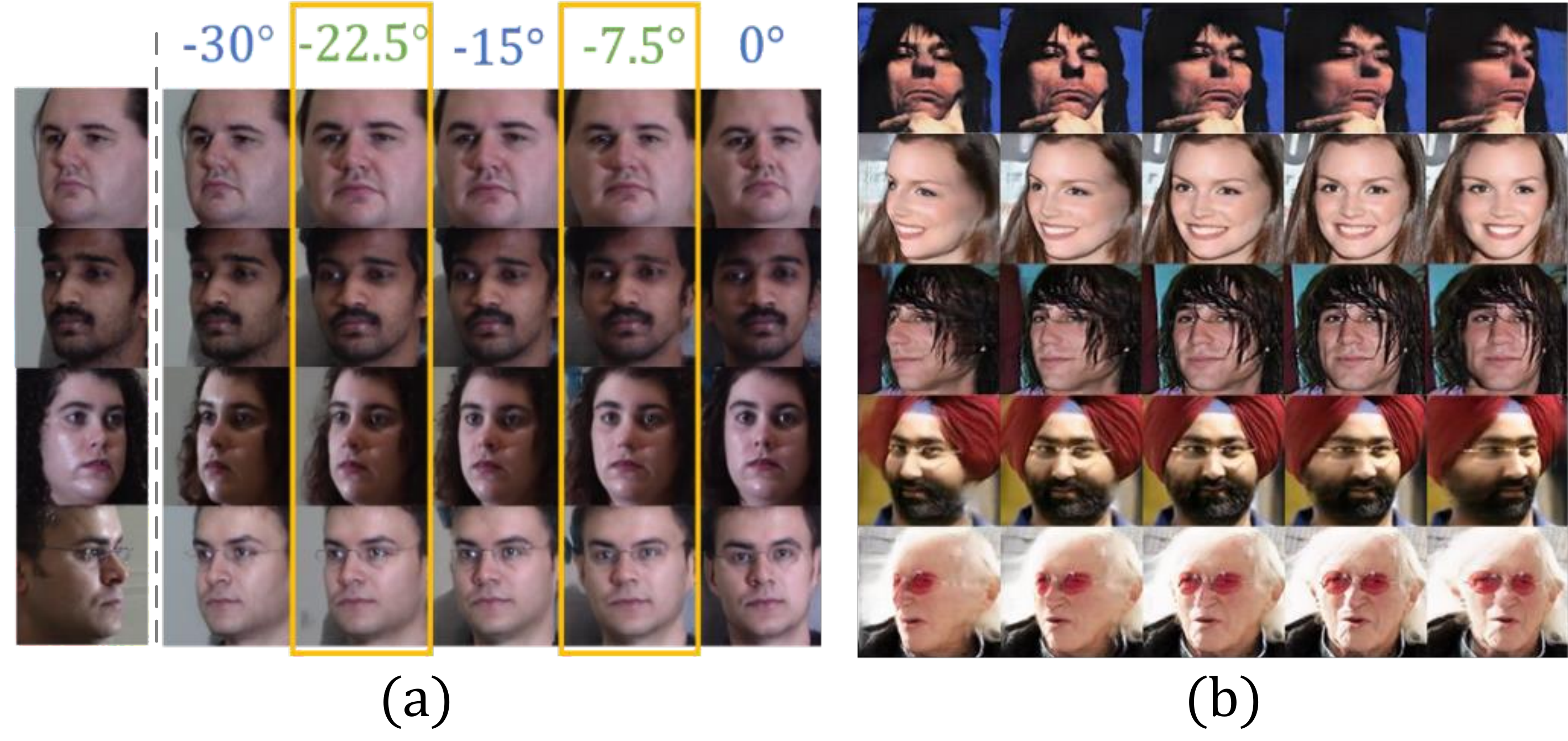}
		\end{center}
        \vspace{-0.5cm}
		\caption{(a) The ID-Unet 
		realizes the translation 
		from the source view to 
		the target, either existing in the MultiPIE dataset ($-30^\circ$,$-15^\circ$,$0^\circ$), 
		or under a new view (inside the yellow box) by the linear interpolation between two adjacent view conditions. (b) Extra results on CelebA from the existing model training on MultiPIE.}
		\label{fig:interpolation}
        \vspace{-0.2cm}
	\end{figure}
	Novel view synthesis, also known as view translation, facilitates the computer to render the same object under arbitrary poses, 	given an input object image in a source pose. 
	This is a challenging task, since it requires the model to understand not only the image content, but also the relation between the object poses and its appearances showing in the image. The model needs to figure out the intrinsic shape of the object and keep it stable during the translation. Meanwhile, it should be able to synthesize the appearance of the object, conforming to the target view condition. 
	
\begin{figure}
		\begin{center} 
			\includegraphics[height=5.4cm]{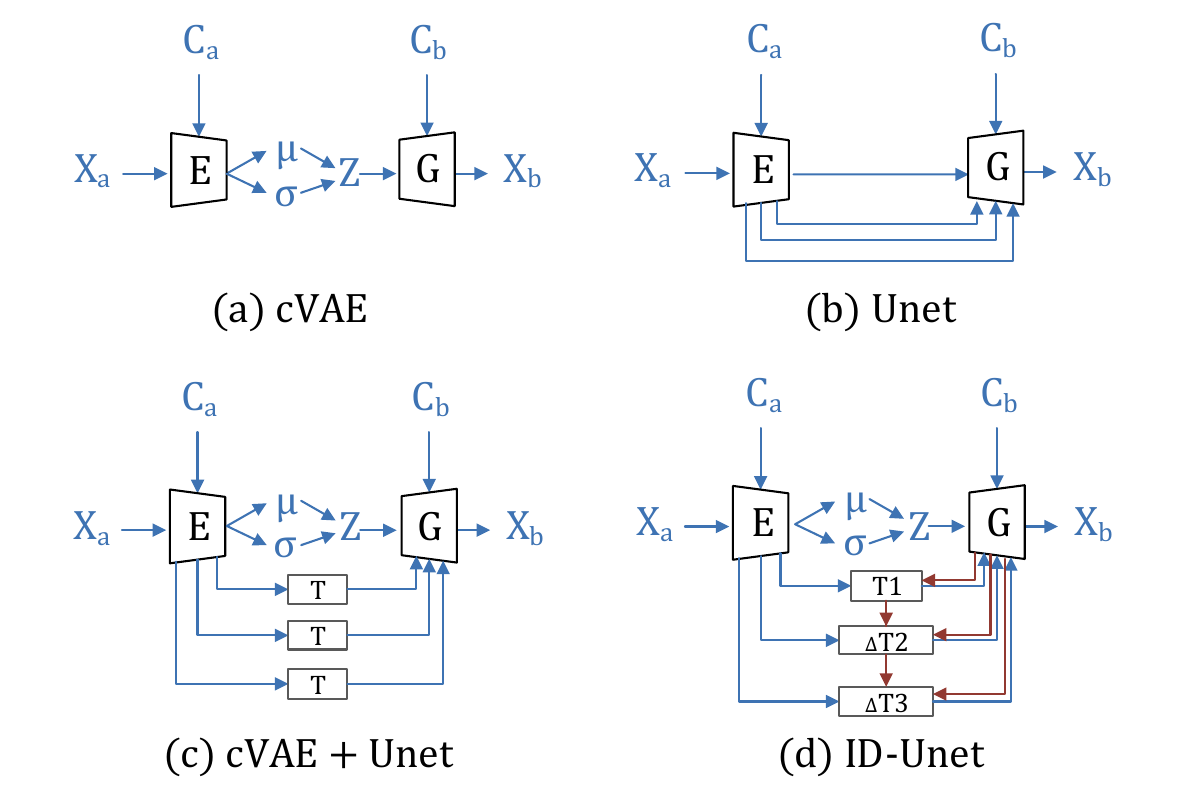}
		\end{center}
        \vspace{-0.3cm}
		\caption{An illustration of several comparing frameworks. (a) and (b) are cVAE and Unet, respectively. (c) is the combination of them, and $\mathrm{T}$ realizes the translation from source view $a$ to target view $b$ based on optical flow. (d) improved from cVAE+Unet, the optical flow is estimated iteratively. The initial flow $\mathrm{T}_1$ is calculated according to the low-resolution features. As the resolution increases layer by layer, the residual $\Delta \mathrm{T}_n$ is calculated to progressively refine the previous result.}
		\label{fig:brief}
        \vspace{-0.3cm}
	\end{figure}

	Recently, learning-based method has been employed broadly for this task.  
	Particularly, 
	view synthesis is commonly regarded as a multi-domain image-to-image translation task, which is often modeled by the autoencoder (AE) \cite{choi2018stargan,xu2019view} or variational autoencoder (VAE) \cite{bao2017cvae,yin2020novel}. Both consist of
	a pair of encoder and decoder, in which only the last layer of the encoder connects to the decoder, as shown in Figure \ref{fig:brief} (a). 
	However, their limitation has already been realized 
	\cite{xiao2018elegant,li2019positional}. Basically, using the latent code from the last layer 
	is not enough to represent the content. 
	Since the decoder can only get one latent code, the source content 
	cannot be kept well in the translated image.  
	A simple but effective solution is the Unet \cite{ronneberger2015u} structure. It utilizes several skipped connections by making the shortcuts from the encoder to the decoder, therefore 
	the output can take more features from the source, as shown in Figure \ref{fig:brief} (b). Such as V-Unet\cite{esser2018variational} is a VAE model with skipped connections and used for person synthesis. Unet indeed improves the image quality. But directly using the low-level encoder features makes it difficult to satisfy the domain requirement, hence the image sometimes fails to be translated into the target domain. 
	
	Intuitively, in view translation, 
	the encoder feature needs to be deformed before giving it to the decoder. A straightforward way is to apply the the same optical flow on the different resolutions of the feature map. The flow can be either determined by the priory knowledge \cite{siarohin2018deformable} or learned by the model \cite{yin2020novel}, and the structure is shown Figure \ref{fig:brief} (c). However, we find that using the same flow on different resolutions limits the model's ability for synthesis. On one hand, the flow is often not accurate enough. It is estimated based on the feature of a certain resolution, therefore may be inappropriate for other sizes. On the other hand, the model can already change the view even without any intentional deformations, which implies that we should give it the flexibility to determine the deformation on different resolutions. 
	
	To properly exploit the encoder features in the view synthesis, this paper proposes an iterative way to deform them in the coarse-to-fine fashion, so that they can be aligned with the corresponding part in the decoder. The deformed features skip several intermediate layers, 
	and are directly given to the layers in the decoder to complement the content details. 
	Inspired by the idea of progressively estimating the optical flow for the raw pixels \cite{lucas1981iterative,baker2004lucas}, our model specifies the offset vectors for the encoder features from the low to the high resolution, and these displacements are accumulated across the multiple resolutions. Specifically, we first use offsets from the low resolution as an approximation to coarsely deform the feature, 
	then the residual offsets are estimated by comparing the roughly deformed result to the decoder feature of the same size. The residuals refine the coarse flow and they are applied to give the additional deformation. The refined flow 
	is further employed by the next block in a larger size. In brief, the encoder feature is first warped according to the coarse flow, and then the remaining offsets is estimated and applied, so that the result is better consistent with the target view.  
	
	To compute the initial flow and its following-up residuals, 
	we design the Soft and Hard Conditional Deformation Modules (SCDM and HCDM) based on the features from the encoder and decoder. The view label is the extra conditional input to control the amount of displacement. The idea of the soft flow is to compute the similarity scores (also known as the attention matrix) between the encoder and decoder features like \cite{vaswani2017attention, wang2018non}. Given the two of them, the spatial and channel similarities 
	are measured, and then applied onto the encoder features to align them into the target view. However, the soft flow is not efficient enough to compute on multiple resolutions. Furthermore, if the target view is far from the source, the similarity may no longer reflect the spatial deformation. Our solution is to estimate the optical flow to "hard" warp the feature before the spatial and channel attention in SCDM. Moreover, we also design the HCDM which gives the 
	high resolution residuals onto 
	the previous small 
	optical flow, and it "hard" warps the current feature and further aligns it to the target view.
	
	The contributions lie in following aspects: (1) We propose an iterative view translation framework which deforms the encoder feature from different layers and gives them to the decoder to improve the synthesis quality. (2) We design the SCDM and HCDM and use them to align the encoder feature into the target view. (3) Extensive experiments on two different datasets show the effectiveness of the proposed framework and our designed modules.
	\section{Related Works}
	\noindent\textbf{GAN and its structure design.} GAN \cite{goodfellow2014generative, odena2017conditional,brock2018large,karras2019style,nguyen2019hologan}
	has shown its ability in synthesizing high dimensional structured data. 
	The rationale behind GANs is to learn the mapping from a latent distribution $z\sim N(0,I)$ to mimic the real data through adversarial training.
	Because of the instability of the adversarial training,
	it often needs to give extra constraints on discriminator $\mathrm{D}$ \cite{gulrajani2017improved,heusel2017gans}. Moreover, by incorporating an encoder $\mathrm{E}$, 
	GAN can be applied in a variety of I2I translation, 
	either supervised by the groundtruth \cite{isola2017image,wang2018high} or not \cite{zhu2017unpaired,choi2018stargan}. 
	In AE, the source image is first converted into a latent code by $\mathrm{E}$, and then $\mathrm{G}$ takes the code and transforms it back into the image. Since there are multiple visual domains, 
	the source and target domain labels are given to the AE as the guide.
	Variational autoencoder (VAE) \cite{kingma2013auto} has the similar structure with AE, 
	in which the latent code is assumed to follow the posterior distribution, 
	and the posterior is to be close to a prior during training. Hence, VAE is not a deterministic model like AE. It can support sampling 
	from the posterior or prior, with their corresponding synthesis looking like real images. 
	VAE is extended to its conditional version cVAE \cite{sohn2015learning,bao2017cvae} as shown in Figure \ref{fig:brief} (a), and cVAE 
	is suitable for either synthesizing the diverse styles of images \cite{zhu2017toward}, or disentangling the latent code \cite{higgins2016beta,zheng2019disentangling}. 
	
	In AE or VAE, $\mathrm{E}$ and $\mathrm{G}$ are only connected through the last latent code, which is not enough to guarantee the synthesis quality. AdaIN \cite{huang2017arbitrary}, SPADE \cite{park2019semantic}, CIN\cite{dumoulin2016learned} and CBIN\cite{nam2018batch} are other ways to inject the feature into the multiple decoder layers through a side branch, which adjusts the statistics of features in the main branch.  The Unet \cite{ronneberger2015u} and its variants link $\mathrm{E}$ and $\mathrm{G}$ by setting up shortcuts between them. But it often leads to failures in I2I translation. Xiao \emph{et al.} \cite{xiao2018elegant} use $\mathrm{G}$'s output as the residual added onto the source image to improve the quality. Li \emph{et al.} \cite{li2019positional} designs PONO 
	layer in Unet, normalizing and adapting source domain features from $\mathrm{E}$ to $\mathrm{G}$. However, these structures are not designed for view synthesis.
		\begin{figure*}[t]
		\begin{center} 
			\includegraphics[height=5.7cm]{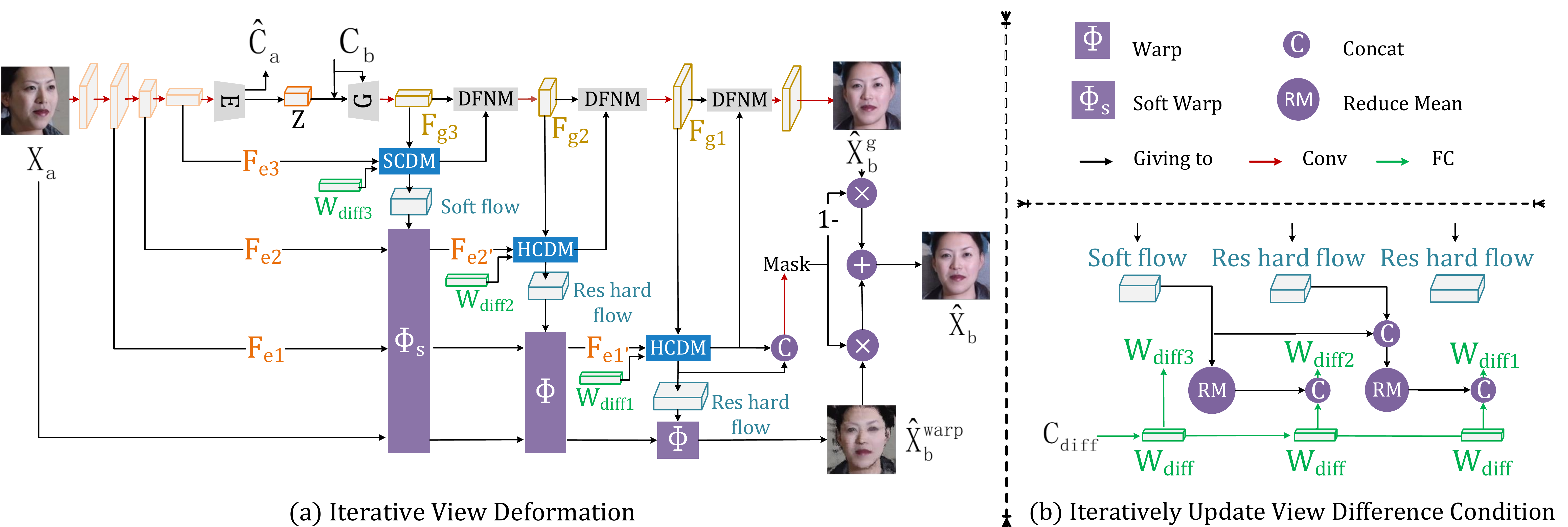}
		\end{center}
        \vspace{-0.3cm}
		\caption{(a) The detailed architecture of the proposed ID-Unet. $X_a$ is mapped to the variable $Z$ through encoder, and it is supplied to the encoder together with the target view label $C_b$. SCDM and HCDM warp the encoder features to the target view, and give their output to the decoder in a way of de-normalization (DFNM) \cite{yin2020novel} to complement the details. And the low resolution optical flow is as an approximation to change the high resolution feature by $\Phi_s$ and $\Phi$, so that the flow is formed in the coarse-to-fine fashion. (b) Iteratively update of the view conditional $C_{diff}$ to $W_{diff3}$, $W_{diff2}$ and $W_{diff1}$.}
		\label{fig:cascade_network_m}
	\end{figure*}

\noindent\textbf{View synthesis.} Traditional approaches \cite{avidan1997novel,kholgade20143d,rematas2016novel} for this task are mainly based on projection geometry, which tries to recover the 3D relation between the object and its projected image plane. They directly estimate either the depth and camera pose \cite{avidan1997novel}, or 3D model parameters \cite{kholgade20143d,rematas2016novel}, so that the object can be projected into the target view.  Learning-based methods \cite{dosovitskiy2015learning,zhou2016view} become increasingly popular nowadays. In \cite{dosovitskiy2015learning}, a CNN model learns to process the latent code for object shape and camera pose, and map it into an image. In \cite{zhou2016view}, the CNN predicts the optical flow to warp the source view into the target. Recently, due to the great success of GAN \cite{park2017transformation,tran2017disentangled,sun2018multi,siarohin2018deformable,xu2019view}, the AE structure plus the adversarial training begins to play the key role in view synthesis. Meanwhile, VAE and its probabilistic latent vector \cite{tian2018cr,yin2020novel} can be applied in this task as well, which even better keeps the contents from the source.  
	However, none of these works consider the coarse-to-fine iterative deformation on features to perform view synthesis. 
	
	

	\section{Method}
	We intend to synthesize object in arbitrary views. Given an image $X_a$ containing an object in the source view $C_a$, and an expected target view $C_b$ as the inputs, the model outputs $\hat{X}_b$, a synthesis of the same object in the target view. The difficulty of this task lies in accurately changing the object from the original to the target view, while keeping other attributes (\emph{e.g.} identity) unchanged during the translation.
	\begin{figure*}
		\begin{center} 
			\includegraphics[height=4.8cm]{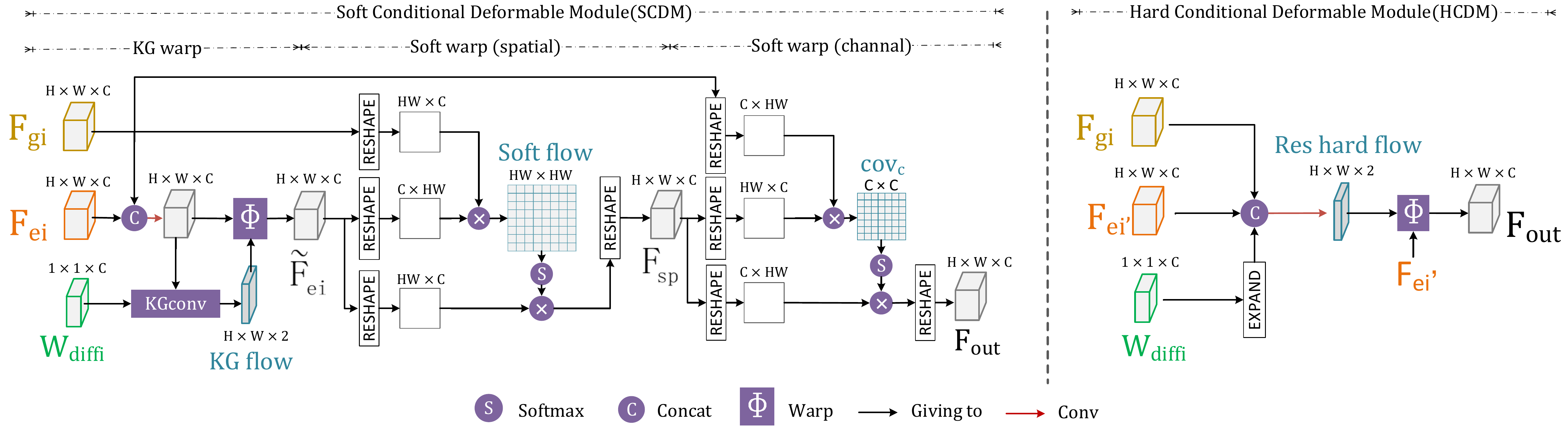}
		\end{center}
		\caption{Illustration of Soft and Hard Conditional Deformation Module. We show the SCDM and HCDM on the left and right respectively. Both have 3 inputs, $F_{gi}$, $F_{ei}$ and $W_{diff}$ from $\mathrm{G}$, $\mathrm{E}$ and view condition, and 1 output $F_{out}$ given to main branch of $\mathrm{G}$. SCDM consists of 3 stages, which are KG warp, spatial soft warp and channel soft warp. HCDM directly estimates the optical flow and warps the encoder feature $F_{ei}$. }
		\label{fig:SCDM&HCDM}
	\end{figure*}
	\subsection{
		The Framework of Iterative View Translation}\label{IVT}
	A brief 
	framework is given in Figure \ref{fig:brief} (d). 
	The idea is to apply multiple deformations on the shallow layer features in the encoder 
	and give them to the decoder, 
	which is conducive to maintain the source content irrelevant to the view. Note that in Figure \ref{fig:brief} (c), module T 
	also estimates the optical flow and is applied on different resolutions, but it is in the independent way. Here the key improvement is the coarse-to-fine manner to estimate the initial deformation $\mathrm{T}_1$ and refine it through $\Delta \mathrm{T}_i$ iteratively, where $i=2,3$ in our setting. 
	Moreover, we find that using the deformed low-level features in the decoder causes the missing of content details in the translated image. 
	While cVAE has a better ability to keep complete objects by introducing the prior distribution as a regularization. The proposed Figure \ref{fig:brief} (d) inherits the cVAE+Unet structure in Figure \ref{fig:brief} (c). In particular, the source view $X_a$ is input to the encoder to provide the content,  
	and is mapped to a posterior distribution,  from which the 
	latent $Z$ can be sampled. The decoder takes $Z$ and the target view condition $C_b$ to synthesize the translation. 

	The proposed ID-Unet, with its specific architecture shown in Figure \ref{fig:cascade_network_m}, accomplishes the iterative view translation on the features $F_e$ in different layers of the encoder, so that they are aligned with their corresponding part $F_g$ in the decoder. 
	Notice that $F_e$ 
	have spatial dimensions and are under the original view $C_a$. If the same 
	features 
	under the target view $C_b$ could also be obtained, it would be easy to 
	estimate the optical flow to deform $F_e$.  
	Intuitively, the decoder feature $F_g$ can be roughly assumed in the target view $C_b$, since the first decoder feature $F_{g3}$ is computed according to the latent $Z$ and condition $C_b$, which has already been aligned into $C_b$. This can be extended to other decoder features such as $F_{g2}$ and $F_{g1}$. They are closer to $C_b$ than their counterparts $F_{e2}$ and $F_{e1}$, so we employ the pair $F_{e3}$ and $F_{g3}$ to estimate the initial $\mathrm{T}_1$, and the following pairs to predict $\Delta \mathrm{T}_i$. 
	
	\subsection{Soft and Hard Deformation}
	We design two types of modules, applying the soft and hard deformations on low and high resolution feature, respectively. Both of them depend on $W_{diff}$, a $1\times 1$ vector given by MLP, which reflects the view difference. We will elaborate it in the next section.\\
	
	\noindent\textbf{Soft Conditional Deformation Module (SCDM)}
	
	SCDM estimates the initial deformation $\mathrm{T}_1$ based on a pair of features $F_{e3}$ and $F_{g3}$ at the lowest resolution, as shown in the left of Figure \ref{fig:SCDM&HCDM}. Instead of directly comparing $F_{e3}$ and $F_{g3}$, a two-channel flow is first predicted through kernel given conv (KGconv) and applied onto $F_{e3}$ by the warping operation $\Phi$. Here, the purpose is to align $F_{e3}$ in the target view direction to form $\tilde{F}_{e3}$, so that the soft flow can be calculated from two similar features $\tilde{F}_{e3}$ and $F_{g3}$, preventing from inappropriate matching two views far from each other. 
	Note that KGconv uses $\text{W}_{diff}$ as conv kernels to generate $x$ and $y$ offsets in the optical flow to assist view translations \cite{yin2020novel}. 
	
	Then, to measure the similarity between source $\tilde{F}_{ei}$ and target $F_{gi}$, we 
	compute 
	the $\text{Soft flow} \in \mathbb{R}^{HW\times HW}$ by the inner product between $\hat{e}_v$ and $\hat{g}_u$:
	$\text{Soft flow}(u, v) = \hat{g}_u^T\hat{e}_v$,
	where $\hat{e}_v$ and $\hat{g}_u \in\mathbb{R}^{C}$ represent the channel-wise centralized feature of $\tilde{F}_{ei}$ and $F_{gi}$ at position $v$ and $u$, $\hat{e}_v = e_v - \mu(e_v)$  and $\hat{g}_u = g_u - \mu(g_u)$.
	$\text{Soft flow}(u)\in \mathbb{R}^{HW}$ represents the similarity between $F_{g3}$ at position $u$ and $\tilde{F}_{e3}$ at all position, so the weighted 
	$\tilde{F}_{ei}$ 
	is the output feature element $F_{sp}(u)$. 
	The weight, $\text{Soft flow}(u)$, is normalized by the Softmax function and multiplied on each position of $\tilde{F}_{ei}$.
	\begin{equation}
	F_{sp}(u) =  \text{softmax} (\frac{1}{\tau} \cdot \text{Soft flow}(u)) \cdot \tilde{F}_{ei}.
	\label{eq:warping}
	\end{equation}
	Different from the classical flow warp (hard warp), 
	$F_{sp}$ in (\ref{eq:warping}) is the weighted sum of %
	the feature at multiple positions in $\tilde{F}_{ei}$. 
	However, smooth weights may change image contents like colors or styles.
	In order to maintain them, 
	we 
	balance the soft and hard warp 
	by incorporating a temperature $\tau<1$ in (\ref{eq:warping}), which increases the impact of the high-weight position (which is more relevant) on the output.
	
	
	Finally, based on $F_{sp}$ and $F_{g3}$,  
	we obtain the similarity matrix $Cov_c$ along the channel in the same way of spatial dimension, and "Soft warp" is also performed on $F_{sp}$ to maintain more valid information in the channel dimension.\\
	
	\noindent\textbf{Hard Conditional Deformation Module (HCDM)}
	
	Basically, HCDM 
	utilizes the results of SCDM, and refines the deformation for larger size $F_{e2}$ and $F_{e3}$. Once the soft flow is obtained,
	the globe deformation $\Phi_s$ can be approximated.
	For the high-resolution features, as shown in Figure \ref{fig:cascade_network_m} (a), 
	$\Phi_s$ also takes effect in HCDM. It first makes the coarse deformation on $F_{e2}$ and $F_{e1}$. Due to the size mismatch between Soft Flow and feature $F_{e2}$ or $F_{e1}$, 
	one element in Soft Flow matrix is scaled 
	and applied to the corresponding 
	square area in the feature of larger size, 
	simplifying as $F_{e2'}=\Phi_s(F_{e2})$. 
	Then the residual optical flow at high resolution is further estimated by 
	the deformed results $F_{e2'}$, the target view features $F_{g2}$ and $W_{diff2}$ together. They are concatenated to learn the residual flow. 
	The residual 
	(Res hard flow) can be superimposed, 
	giving $F_{e1'}=\Phi(\Phi_s(F_{e1}))$, in which $\Phi$ denotes the hard warping operation by the optical flow. Therefore, with the increase on 
	resolution, 
	the optical flow 
	for translation is gradually refined by HCDM.

	\subsection{Iteratively Update View Difference Condition}
	With the gradual refinement of optical flow, the features $F_{e2'}$ and $F_{e1'}$ 
	have been converted to the target view to a certain extent. Then 
	the actual view 
	of the current features ($F_{e2'}$ or $F_{e1'}$) is no longer the same as the source, and the 
	condition $W_{diff}$ should also be adapted, %
	since it no longer translates from the source to the target, but from the current view to the target. In our model, 
	$W_{diff}$ is updated iteratively together with the feature. 
	Specifically, we use the current 
	flow to measure the amount of 
	the translation, and learn 
	how to update $W_{diff}$ by the model itself.
	In Figure \ref{fig:cascade_network_m} (b), the view label difference 
	$C_{diff}$ is passed through an MLP, 
	to get $W_{diff}$. 
	$W_{diff3}$ used for the first warp is directly obtained from $W_{diff}$ through one fc layer. During the further operation, the mean of optical flow $(\mu(\mathrm{d}x),\mu(\mathrm{d}y))$ 
	is concatenated with $W_{diff}$ to determine the next conditional vector ($W_{diff2}$ or $W_{diff1}$) for the further deformation.
	\subsection{Training Details and Loss Functions}
	\noindent\textbf{Adversarial and Reconstruction Loss}
	
	We use adversarial loss $L_{\mathrm{E}, \mathrm{G}}^{adv}$ and $L_{\mathrm{D}}^{adv}$ \cite{lim2017geometric} to 
	ensure the 
	translated image approximates the true distribution like in (\ref{eq:adv DEG}). As shown in Figure \ref{fig:cascade_network_m} (a), the 
	final $\hat{X}_b$ is 
	mixed 
	by two parts. One is the 
	$\hat{X}^{warp}_b$, obtained 
	by the soft and hard 
	deformation 
	on the source 
	$X_a$, and the other $\hat{X}^g_b$ is the output of the generator. The model learns a single channel mask to weight and combine the two results. The mask is computed based on the output and the optical flow in the last HCDM. 
	\begin{equation}
	\label{eq:adv DEG}
	\begin{aligned}
	L_{\mathrm{D}}^{adv}  =& \mathbb{E}_{{X}}[\max(0,1-\mathrm{D}(X,C_b))]\\
	+& \mathbb{E}_{\hat{X}_b}[\max(0,1+\mathrm{D}(\hat{X}_b,C_b))],\\ 
	L_{\mathrm{E}, \mathrm{G}}^{adv} =&\mathbb{E}_{\hat{X}_b}[\max(0,1-\mathrm{D}(\hat{X}_b,C_b))]
	\end{aligned}
	\end{equation}
	Like ACGAN \cite{odena2017conditional}, we use classification losses $L_\mathrm{C}^{cls}$ and $L_{\mathrm{E}, \mathrm{G}}^{cls}$ in (\ref{eq:cls EGD}). The classifier $\mathrm{C}$ shares a part of its weights with discriminator $\mathrm{D}$.
	\begin{equation}
	\label{eq:cls EGD}
	\begin{aligned}
	L_{\mathrm{C}}^{cls}  = - \mathbb{E}_{X_b}\sum_c \mathbb{I}(c=C_b) \log \mathrm{C}(c|X_b ),\\
	L_{\mathrm{E}, \mathrm{G}}^{cls} =- \mathbb{E}_{\hat{X}_b}\sum_c \mathbb{I}(c=C_b) \log \mathrm{C}(c|\hat{X}_b )
	\end{aligned}
	\end{equation}
	In addition, 
	by combining the reconstruction loss in image domain $L_{\mathrm{E}, \mathrm{G}}^{pixel} = ||X -\hat{X}_j||_1$ and feature domain $L_{\mathrm{E}, \mathrm{G}}^{content}= \sum_i ||\phi^i({X}) -\phi^i({\hat{X}_j})||_1$, the image 
	quality is guaranteed more faithfully.
	Here $\phi$ indicates $i$-th layer of a pre-trained VGG \cite{simonyan2014very} network, and $j=b,a,aa$. $\hat{X}_{a}$ and $\hat{X}_{b}$ are the fake images at target view A and B. 
	$\hat{X}_{aa}$ 
	the cyclic translation result, which is translated back from the synthesised image in view B.
	\begin{figure}
		\begin{center} 
			\includegraphics[height=3.0cm]{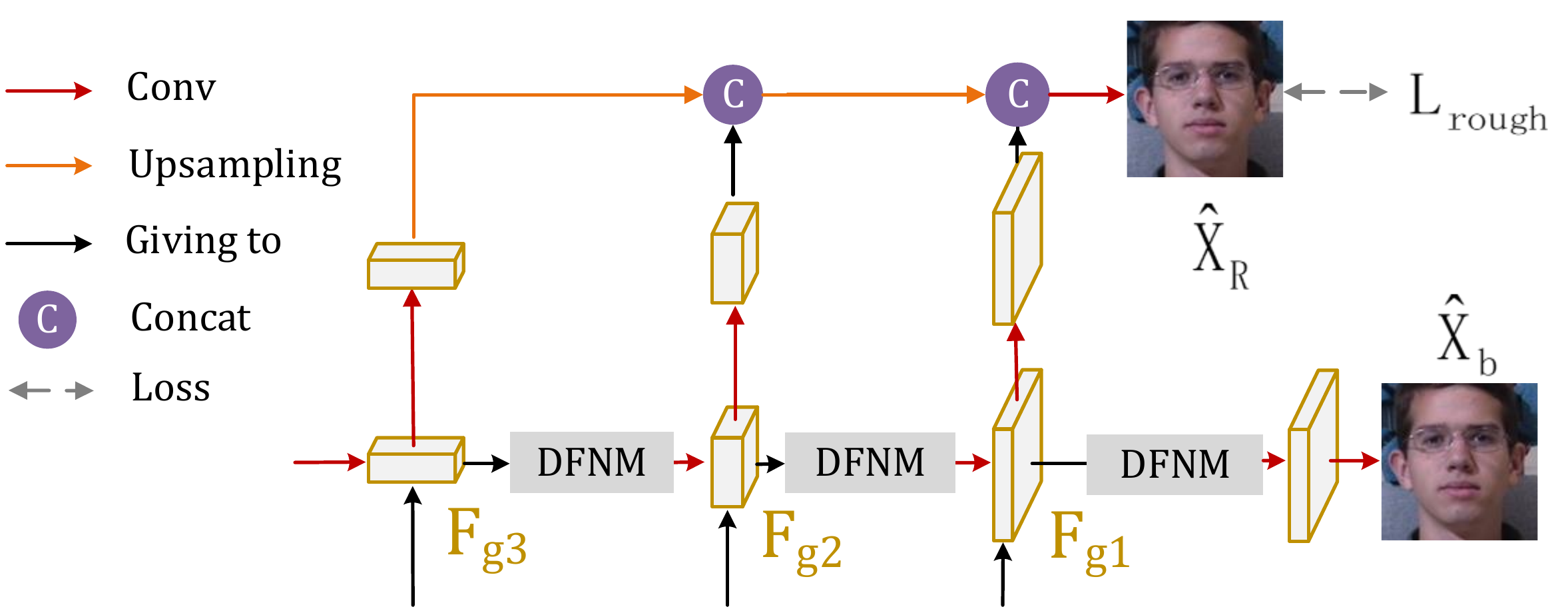}
		\end{center}
		\caption{Besides the normal translated image $\hat{X}_b$, image $\hat{X}_R$ is generated from $F_{g3}$, $F_{g2}$ and $F_{g1}$ for the rough loss.}
		\label{fig:rough_loss}
	\end{figure}
	
	\noindent\textbf{Disentangling Loss}
	
	The source image $X_a$ is mapped to a code $Z\sim \mathrm{E}(Z|X_a)$ where $\mathrm{E}(Z|X_a)$ is a posterior depending on the source $X_a$. $Z$ is fed directly into $\mathrm{G}$, so it should keep the content of the object, and be irrelevant to views \cite{xu2019view,yin2020novel}. To prevent $Z$ from taking view relevant factors, we add two auxiliary classifier losses for $\mathrm{E}$. One computes the classification loss $L_E^{clsC}$ which 
	tries to predict $\hat{C}_a=\mathrm{E}(c|X)$ to approximate view label $C_a$, 
	as is defined in the first term in (\ref{eq:advE}). 
	Another 
	adversarial constraint $L_{\mathrm{E}}^{cls}$ in (\ref{eq:advE}) makes the view classification 
	based on $Z$ by the hidden layer classifier $\mathrm{DAC}$, which is the last two terms in (\ref{eq:advE}). 
	\begin{equation}
	\label{eq:advE}
	\begin{aligned}
	L_{\mathrm{E}}^{clsC} = &- \mathbb{E}_{X\sim X_a}\sum_c \mathbb{I}(c=C_a) \log \mathrm{E}(c | X),\\
	L_{\mathrm{E}}^{clsZ} =& - \mathbb{E}_{Z\sim \mathrm{E}(Z|X_a)}\sum_c \frac{1}{C} \log \mathrm{DAC}(c|Z),\\
	L_{\mathrm{DAC}}^{clsZ} =& - \mathbb{E}_{Z\sim \mathrm{E}(Z|X_a)}\sum_c \mathbb{I}(c=C_a) \log \mathrm{DAC}(c|Z)
	\end{aligned}
	\end{equation}
	Here $L_{\mathrm{DAC}}^{clsZ}$ is the penalty to train $\mathrm{DAC}$, ensuring the accuracy of the view classification. 
	$L_{\mathrm{E}}^{clsZ}$ is the adversarial loss applied on $\mathrm{E}$ to make $\mathrm{DAC}$ confused
	to predict the uniform value on each view. Furthermore, via the constraint of KL loss 
	$L_{KL} =  D_\text{KL}[\mathrm{E}(Z|X_a)||{N}({0},{ I})]$, 
	the latent code $Z$ from the encoder is close to the standard normal distribution and has no category-related information. 
	
	\noindent\textbf{Rough Loss}
	
	We design the rough loss 
	on the deformed features 
	in SCDM and HCDM, to make the features conform to the target view. 
	As is described in section \ref{IVT} and  Figure \ref{fig:cascade_network_m}, 
	the decoder features $F_{g3}$, $F_{g2}$ and $F_{g1}$ are assumed under target view $C_b$.
	To better ensure 
	that they are in target view, 
	$F_{g3}$, $F_{g2}$ and $F_{g1}$ are combined and fed to a layer $\psi$ to generate an image $\hat{X}_R = \psi (F_{g3}, F_{g2}, F_{g1})$ as shown in Figure \ref{fig:rough_loss}. 
	The image $\hat{X}_R$ is constrained 
	by pixel-wise L1 loss and classification loss of the classifier $\mathrm{C}$, like in (\ref{eq:rough}).\\
	\begin{equation}
	\label{eq:rough}
	L^{rough}_{\mathrm{E}, \mathrm{G}} = || X_b -\hat{X}_R||_1 + \sum_c \mathbb{I}(c=C_b) \log \mathrm{C}(\hat{X}_{R})
	\end{equation}
	\textbf{Overall Objective.}
	The total optimization loss is a weighted sum
	of the above. Generators $\mathrm{E},\mathrm{G}$, discriminator $\mathrm{D}$, classifier $\mathrm{C}$, and the latent classifier $\mathrm{DAC}$ are
	trained by minimizing (\ref{eq:E all}).
	\begin{equation}
	\label{eq:E all}
	\begin{aligned}
	L_{\mathrm{E}, \mathrm{G}} =& L_{\mathrm{E}, \mathrm{G}}^{adv} + L_{\mathrm{E}, \mathrm{G}}^{cls} + \alpha_1 L_{\mathrm{E}, \mathrm{G}}^{content} + \alpha_2L_{\mathrm{E}, \mathrm{G}}^{pixel}\\
	+& \alpha_3L_{KL} +  L_{\mathrm{E}}^{clsC} +  L_{\mathrm{E}}^{clsZ} + \alpha_4L^{rough}_{\mathrm{E}, \mathrm{G}},\\
	L_{\mathrm{D}} =& L_{\mathrm{D}}^{adv},\quad L_{\mathrm{C}} =  L_{\mathrm{C}}^{cls},\quad
	L_{\mathrm{DAC}} =  L_{\mathrm{DAC}}^{clsZ}
	\end{aligned}
	\end{equation}
    The loss weights $\alpha_1$, $\alpha_2$, $\alpha_3$, $\alpha_4$ = 5, 5, 0.1, 10. 
	
	\begin{table*}[ht] 
		\begin{center}
			\setlength{\tabcolsep}{1.5mm}{
				\begin{tabular}{ l l l c c c c c c c c c c}
					\toprule
					& Method && \multicolumn{5}{c}{MultiPIE}& &\multicolumn{4}{c}{3D chair}  \\
					\cline{4-8}\cline{10-13}
					&&&$L_1\downarrow$ &SSIM$\uparrow$ &LPIPS$\downarrow$&FID$\downarrow$&id-acc$\uparrow$&  &$L_1\downarrow$&SSIM$\uparrow$&LPIPS$\downarrow$&FID$\downarrow$ \\
					\midrule
					& MV\cite{sun2018multi} & & $15.21$  & $0.489$ &$0.217$& $29.85$ & 0.742 && $13.86$ &$0.779$  & $0.224$ &$104.49$  \\
					& Unet\cite{ronneberger2015u} & & $14.03$  & $0.619$ &$0.164$& $49.86$ & 0.396 && $21.75$ &$0.697$  & $0.255$ &$86.74$  \\				
					& cVAE\cite{bao2017cvae} & & $12.82$  & $0.635$   &$0.119$& $28.99$  &0.651 && $8.93$ &$0.828$ & $0.102$ &$27.79$   \\				 	
					& CRGAN\cite{tian2018cr} &  & $14.12$ & $0.627$ &$0.141$ &$26.77$  & $0.868$  && $13.33$ &$0.788$   & $0.196$ &$28.23$  \\				
					& VIGAN\cite{xu2019view} &  &  $12.96$  &  $0.638$ &$0.117$&  $29.05$ & $0.686$ &&  $12.13$ &$0.781$  & $0.133$ &$33.18$  \\
					& PONO\cite{li2019positional} &  &  $13.63$  &  $0.621$ &$0.126$&  \boldmath{$23.77$} & $0.862$ &&  $12.74$ &$0.780$  & $0.148$ &$37.85$  \\
					& CDVAE\cite{yin2020novel} &  &  $13.49$  &  $0.623$ &$0.125$&  $23.95$ & $0.917$ &&  $13.38$ &$0.773$  & $0.148$ &$40.81$  \\
					\midrule
					& cVAE+Unet &  & $12.37$ & $0.658$ &$0.113$ &$28.98$  & $0.689$ && $11.32$ &$0.790$   & $0.123$ &$32.04$  \\				
					& A:cVAE+Unet+Iterative &  &  $12.14$  &  $0.676$ &$0.100$&  $27.50$ & $0.893$ &&  $10.64$ &$0.801$  & $0.120$ &\boldmath{$27.76$}  \\			
					& B:A+ImageMix & & $12.01$  & $0.679$ &$0.101$& $26.55$ & \boldmath{$0.928$} && $9.30$ &$0.819$  & $0.104$ &$36.18$ \\				
					& C:B+IterativeC & & $11.11$  & $0.684$   &$0.095$& {$24.55$}  &0.913 && $9.055$ &$0.826$  & $0.102$ &$29.06$  \\				
					& D:C+rough loss & & \boldmath{$10.72$}  & \boldmath{$0.694$}   &\boldmath{$0.093$} &  {$25.12$}  & {$0.911$} && \boldmath{$7.57$} & \boldmath{$0.847$}  & \boldmath{$0.089$} &$28.87$  \\		
					\bottomrule
				\end{tabular}
			}
		\end{center}
        \vspace{-0.3cm}
		\caption{Comparison on the MultiPIE and the 3D chair datasets. 
		}
		\vspace{-0.4cm}
		\label{Table1}
	\end{table*} 

	\section{Experiments}
	\subsection{Datasets and Quantitative Metrics.}
	\textbf{Datasets.} 
	We validate the proposed ID-Unet on face dataset MultiPIE\cite{gross2010multi}
	and 3D chair\cite{aubry2014seeing}
	object dataset. MultiPIE contains about 130,000 images, with 13 viewing angles, spanning $180^\circ$.
	Nine of central viewing angles are used for training and testing.
	The 3D chair contains 86,304 images,
	covering a total of 62 angles. 
	For all the datasets, 80\% are used for training and the rest 20\% for testing.
	
	\textbf{Quantitative Metrics. }
    To give the 
	evaluation on different methods, we use following metrics during the test.
	We calculate \textbf{L1 error} and \textbf{LPIPS} \cite{zhang2018unreasonable} to measure the difference at pixel level and feature level between the generated and ground truth image. \textbf{SSIM} \cite{wang2004image} is calculated to compare the similarity of image structure. \textbf{FID} represents the distance between the generated image distribution and the real image distribution, so as to measure the authenticity of the generated image. At the same time, on the MultiPIE dataset \cite{gross2010multi}, we use the face identity recognition network pretrained on VGGface \cite{parkhi2015deep} dataset to calculate the \textbf{identity accuracy} of generated image. Table \ref{Table1} lists all the metrics for the ablation and comparison models. More specific training details are given in the supplementary materials.
	\subsection{Ablation Study}
	In this section, we compare the results 
	in several different ablation settings to verify the effectiveness of every component in the proposed method. 
	
\begin{figure}[t]
		\centering
		
		\subfigure{
			\begin{minipage}[t]{0.51\linewidth}
				\centering
				\includegraphics[width=1.7in]{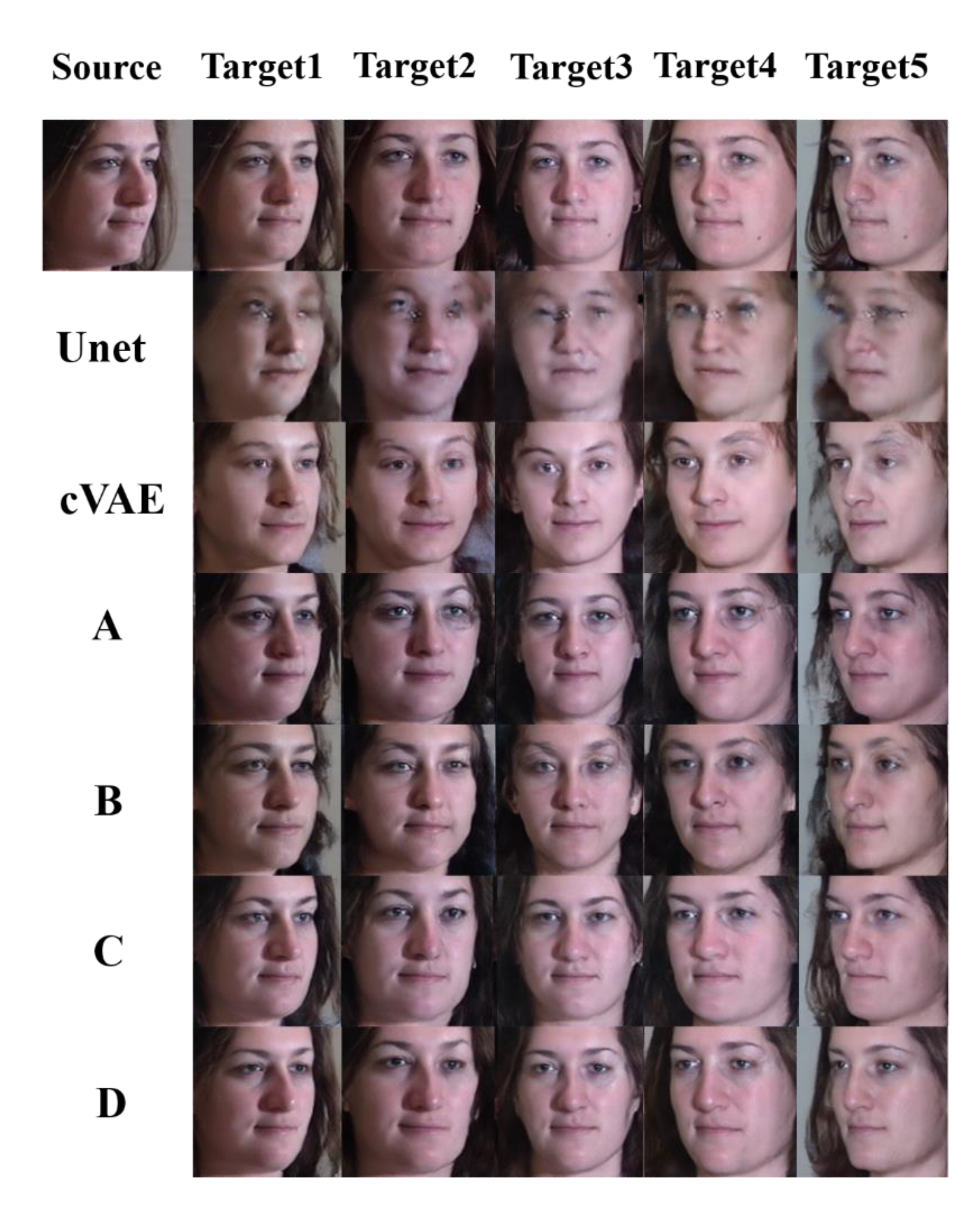}\\
				\includegraphics[width=1.7in]{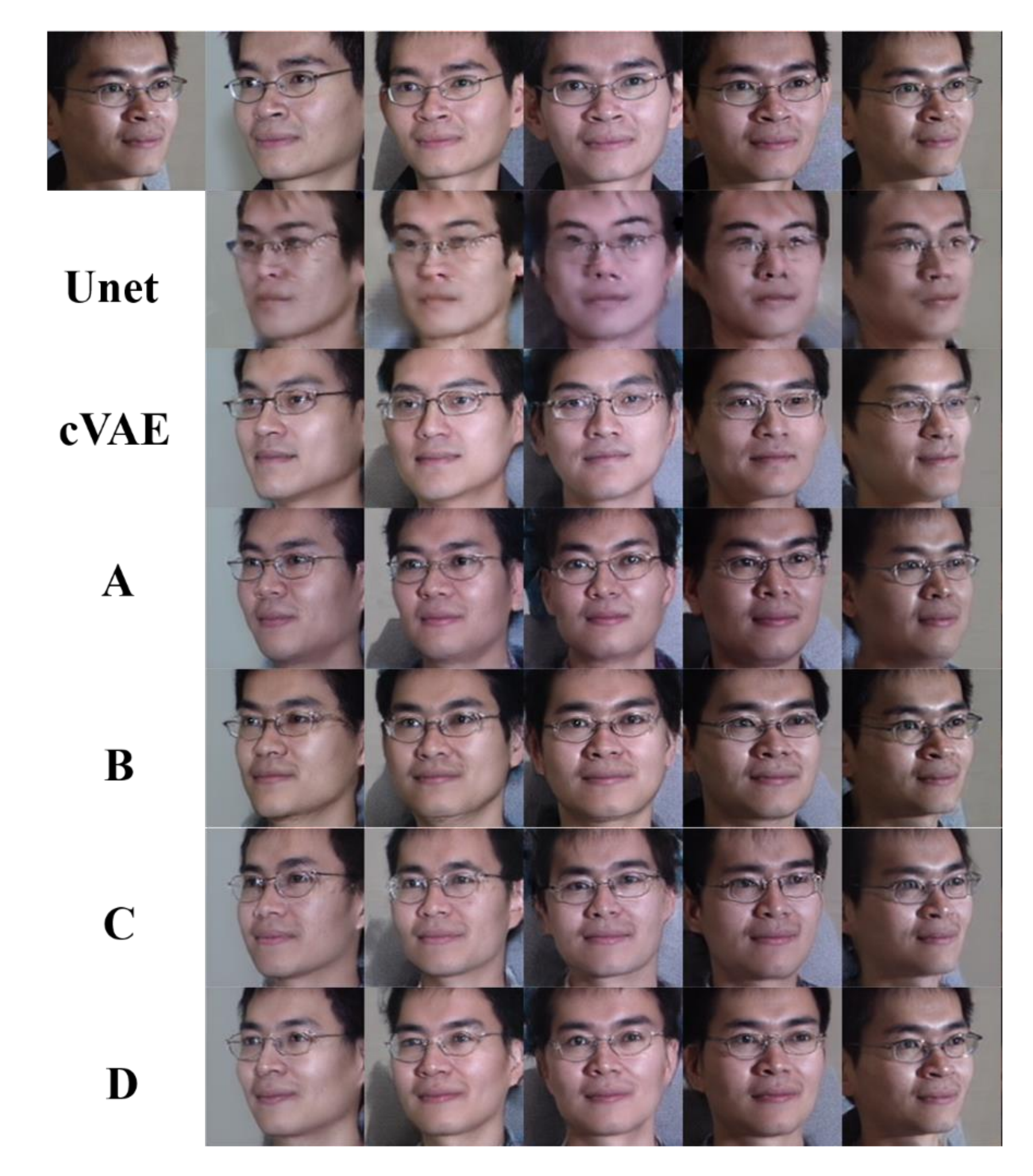}\\ 
			\end{minipage}%
		}%
		\hspace{-0.18in}
		\subfigure{
			\begin{minipage}[t]{0.49\linewidth}
				\centering
				\includegraphics[width=1.7in]{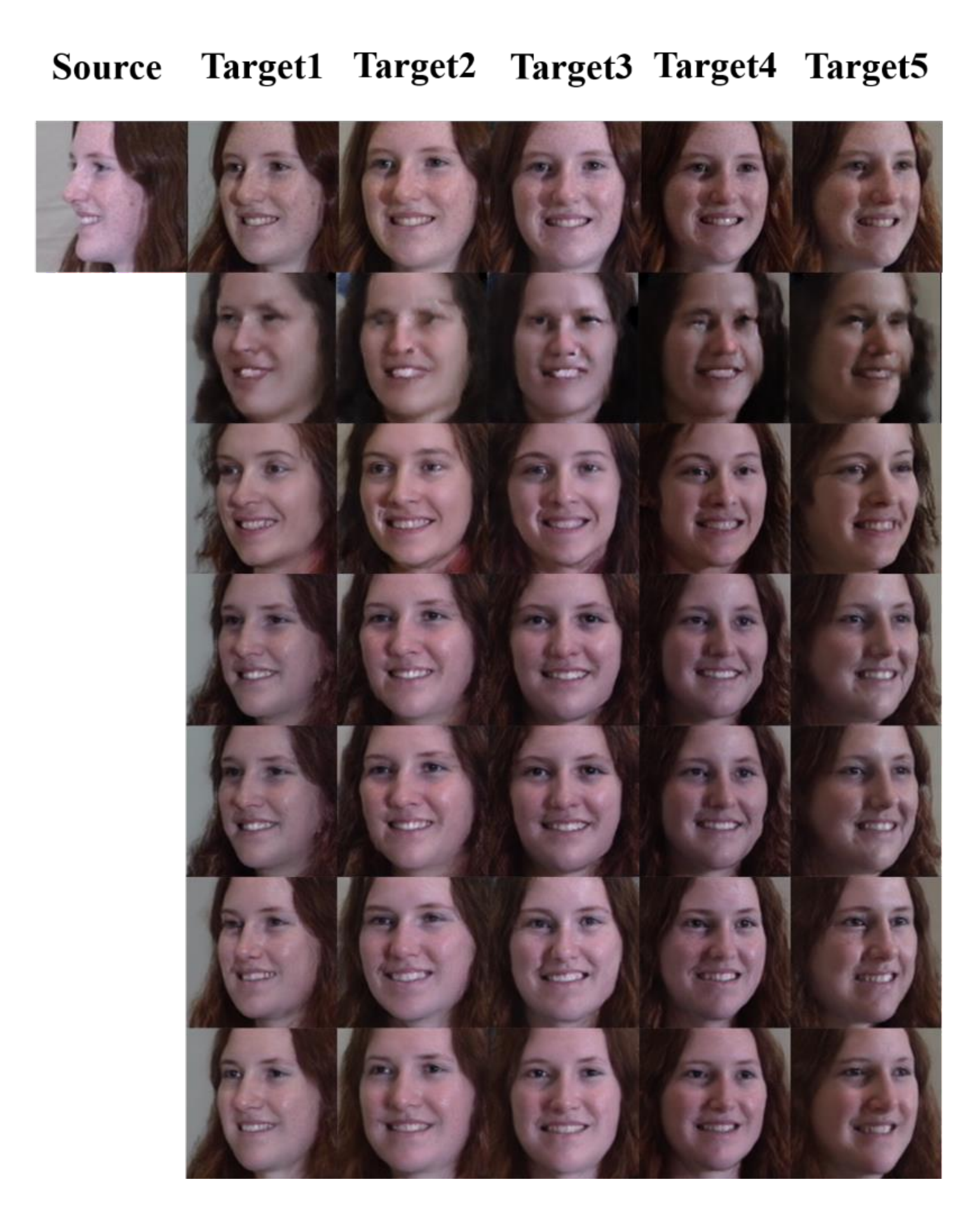}\\ 
				\includegraphics[width=1.7in]{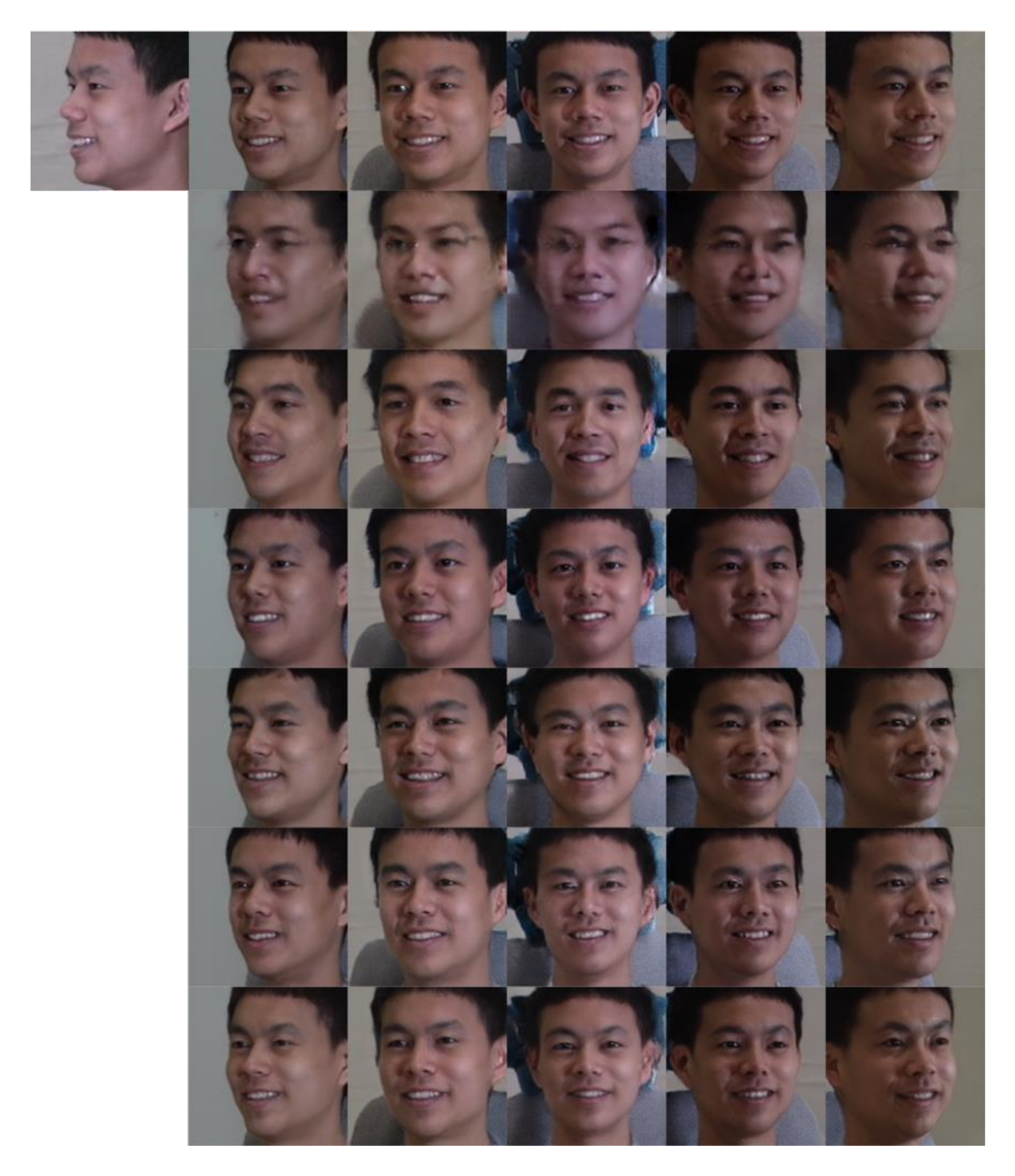}\\ 
			\end{minipage}%
		}%
		\centering
        \vspace{-0.3cm}
		\caption{Ablation study on MultiPIE dataset. The source and the ground truth targets are provided in the first row. Please zoom in for details.} 
		\label{fig:ablation_multiPIE}
        \vspace{-0.6cm}
	\end{figure}

\begin{figure}
		\begin{center} 
			\includegraphics[height=8.5cm]{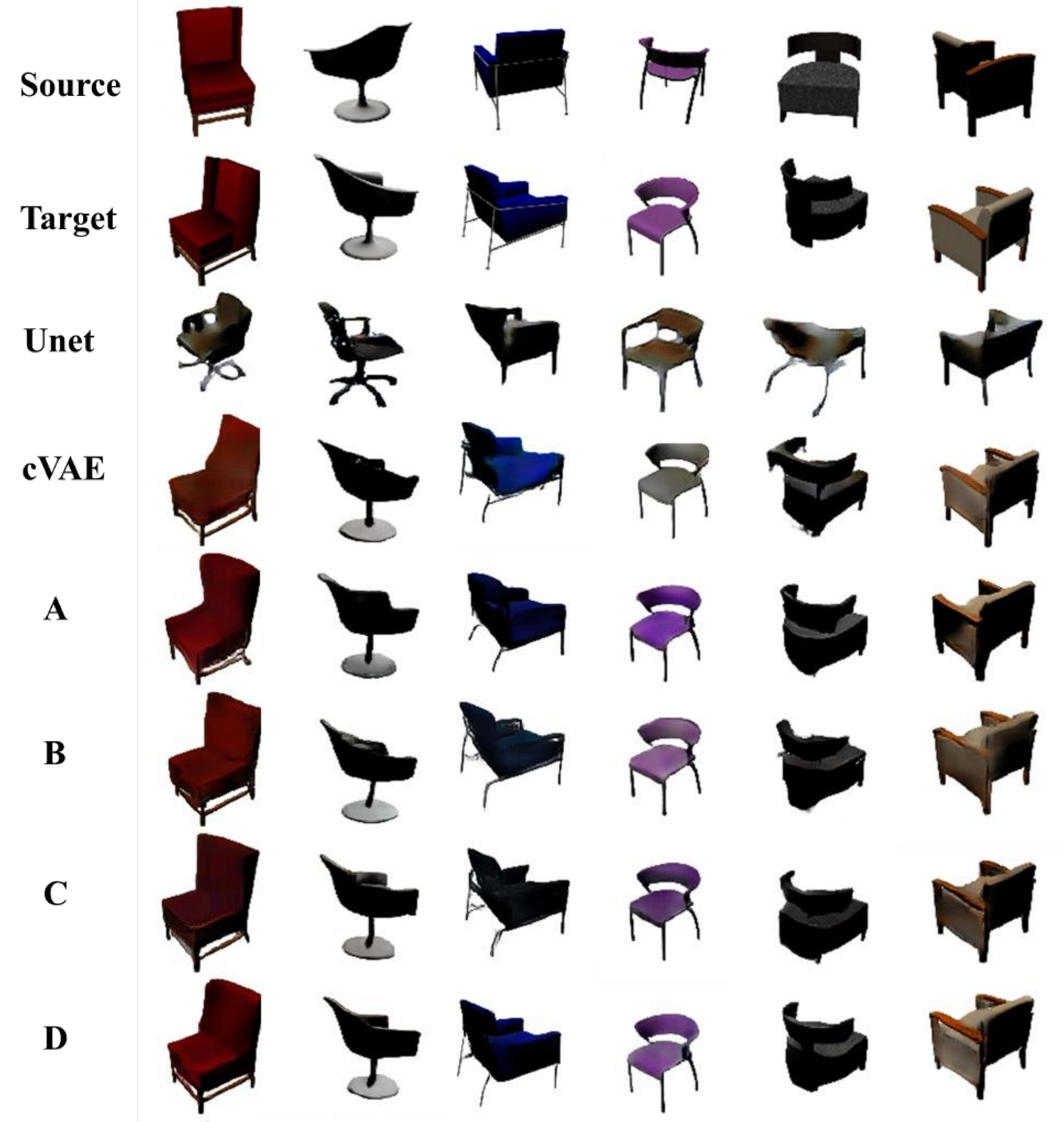}
		\end{center}
        \vspace{-0.3cm}
		\caption{Ablation study on 3D chair dataset. The source and ground truth targets are given in the 1st and 2nd rows.}
		\label{fig:ablation_chair}
        \vspace{-0.3cm}
	\end{figure}

	\textbf{A: cVAE+Unet+Iterative.} 
	Setting A is based on the 
	two common 
	models Unet and cVAE, combining them and then sending the encoder features 
	to the corresponding decoder layer after iterative view translation. 
	In Figure \ref{fig:ablation_multiPIE} and \ref{fig:ablation_chair}, 
	the 2nd, 3rd and 4th rows are generated images from Unet, cVAE and model A, respectively. 
	We observe that the object from Unet appears incomplete (disappeared chair part or eyes). For cVAE, 
	the face identity 
	and the chair color have changed to a large extent. 
	While the setting A can ensure the integrity of the image and the invariance of 
	the information irrelevant to the view. 
	Meanwhile, as shown in Table \ref{Table1}, compared with Unet and cVAE, all results under setting A are significantly improved, especially the id-acc increases from 0.396 (Unet), 0.651 (cVAE) to 0.893. 
	
	\textbf{B: A+ImageMix.} Based on A, setting B combines the output of the generator $\hat{X}^g_b$ with the deformation of the original image $\hat{X}^{warp}_b$, which is conducive to maintain more valid content of the original image and generating more realistic images, as shown in the 5th row in Figure \ref{fig:ablation_multiPIE}, with the id-acc reaching 0.928. 
	
	\textbf{C: B+IterativeC.} The experimental setting C further extends on B. In Figure \ref{fig:ablation_multiPIE} and \ref{fig:ablation_chair}, the view translation 
	is more accurate and better handled in detail. Because the view difference condition $W_{diffi}$, where $i = 1,2,3$, is updated iteratively according to the degree of deformation of current features, the view 
	condition is better adjusted and controlled. The result in Table \ref{Table1} also verifies the conclusion.
	
	\textbf{D: C+rough loss.} 
	In setting D, the effectiveness of rough loss is validated. From the 
	last row in Figure \ref{fig:ablation_chair}, it can be seen that the chairs are not only close to the targets on pixel, but also have stable shape at different views. It is obvious that this model can better understand the intrinsic shape of the chairs.
	This is also supported by Table \ref{Table1}.
 
	\begin{figure}
	    \begin{center}   
			\includegraphics[height=8.0cm]{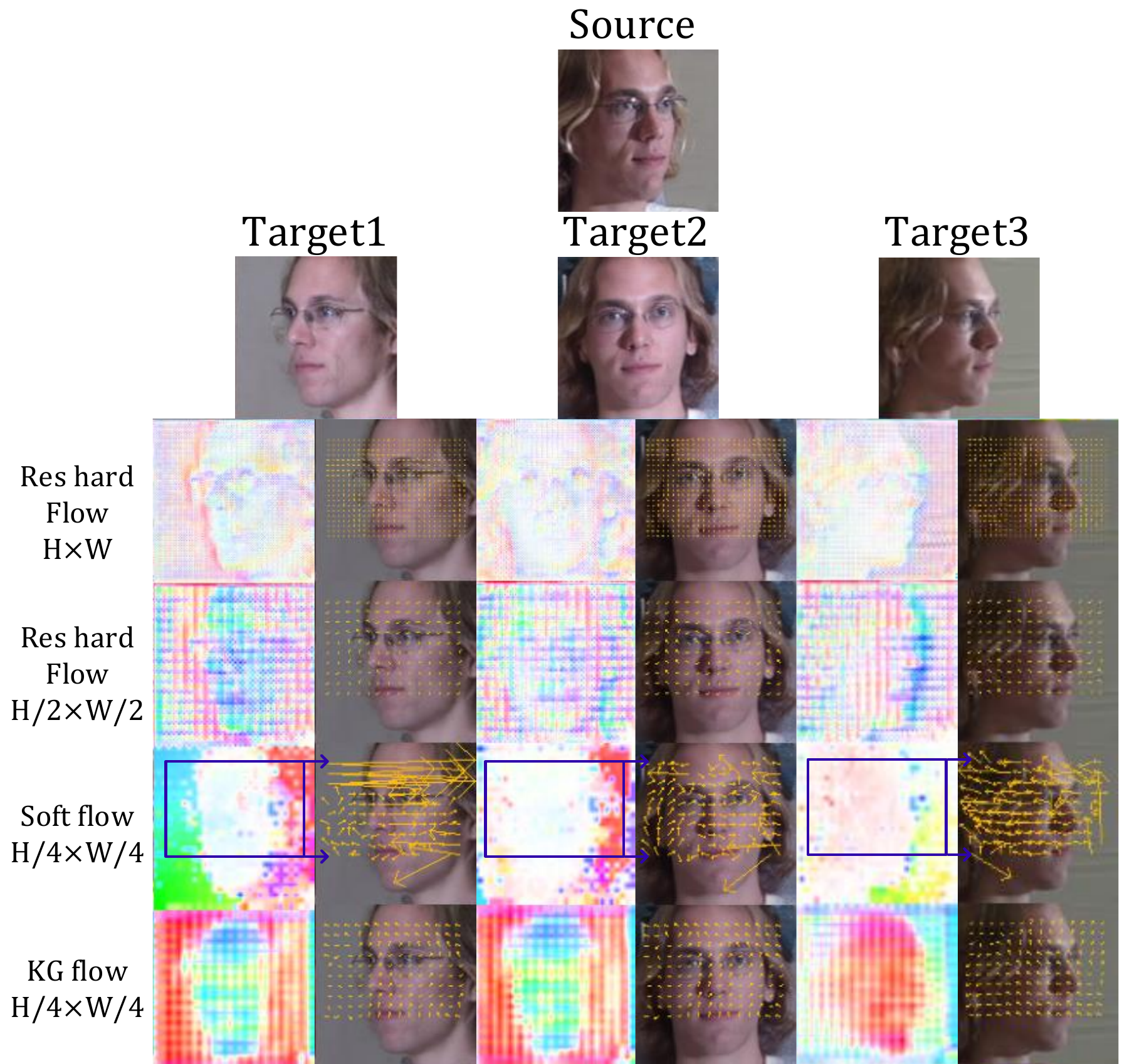} 
		\end{center}
        \vspace{-0.3cm}
		\caption{Visualization of optical flow on different layers. We list 4 deformation flows from the bottom to the top. 
		The direction of the flow points from the target to source.}
		\label{fig:flow_multiPIE}
	\end{figure}

	\begin{figure*}[h]
		\centering
		\begin{minipage}{0.321\textwidth}
			\centering
			\includegraphics[width=2.35in]{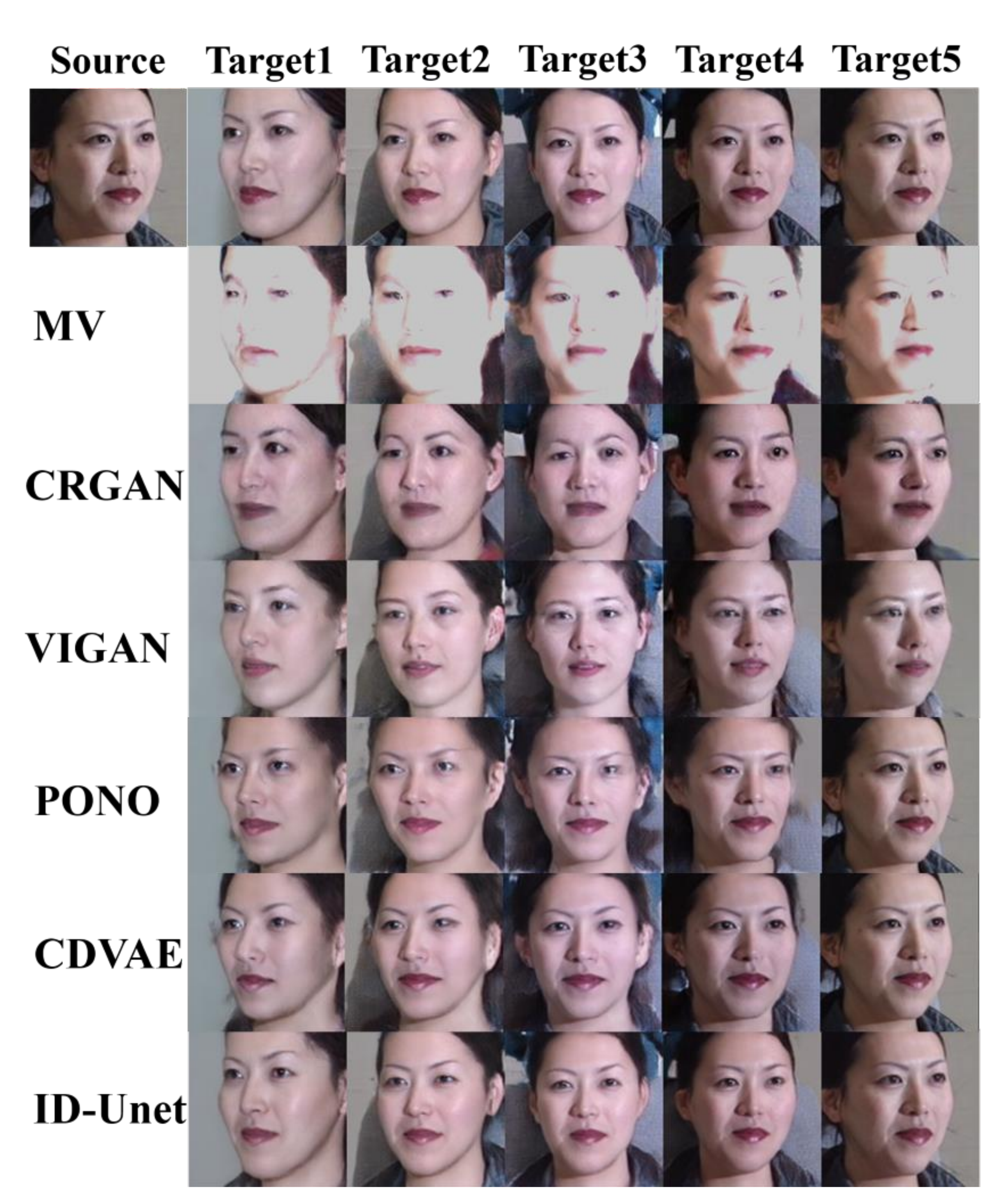}
		\end{minipage}
		\begin{minipage}{0.320\textwidth}
			\centering
			\includegraphics[width=2.36in]{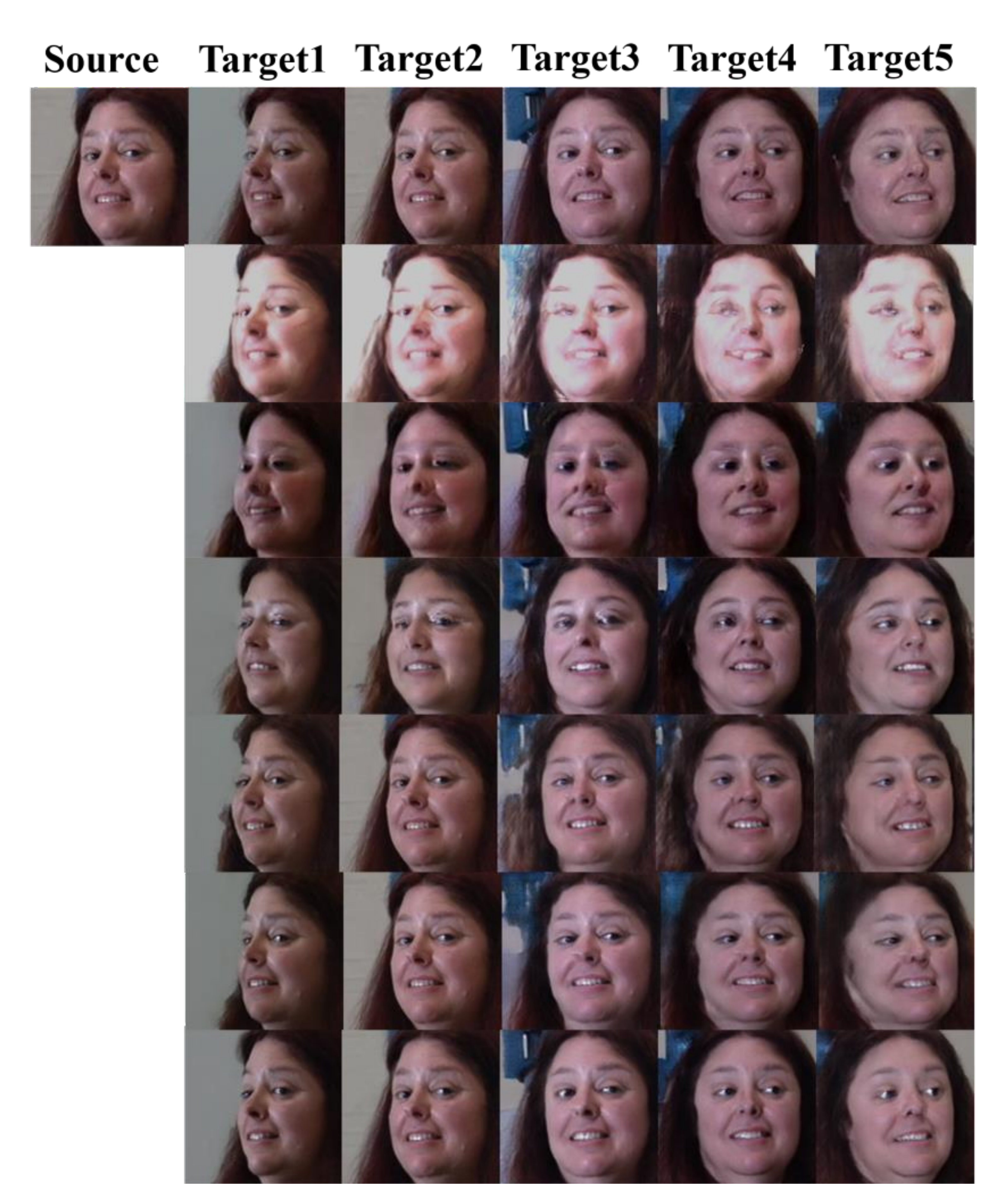}
		\end{minipage}
		\begin{minipage}{0.327\textwidth}
			\centering
			\includegraphics[width=2.34in]{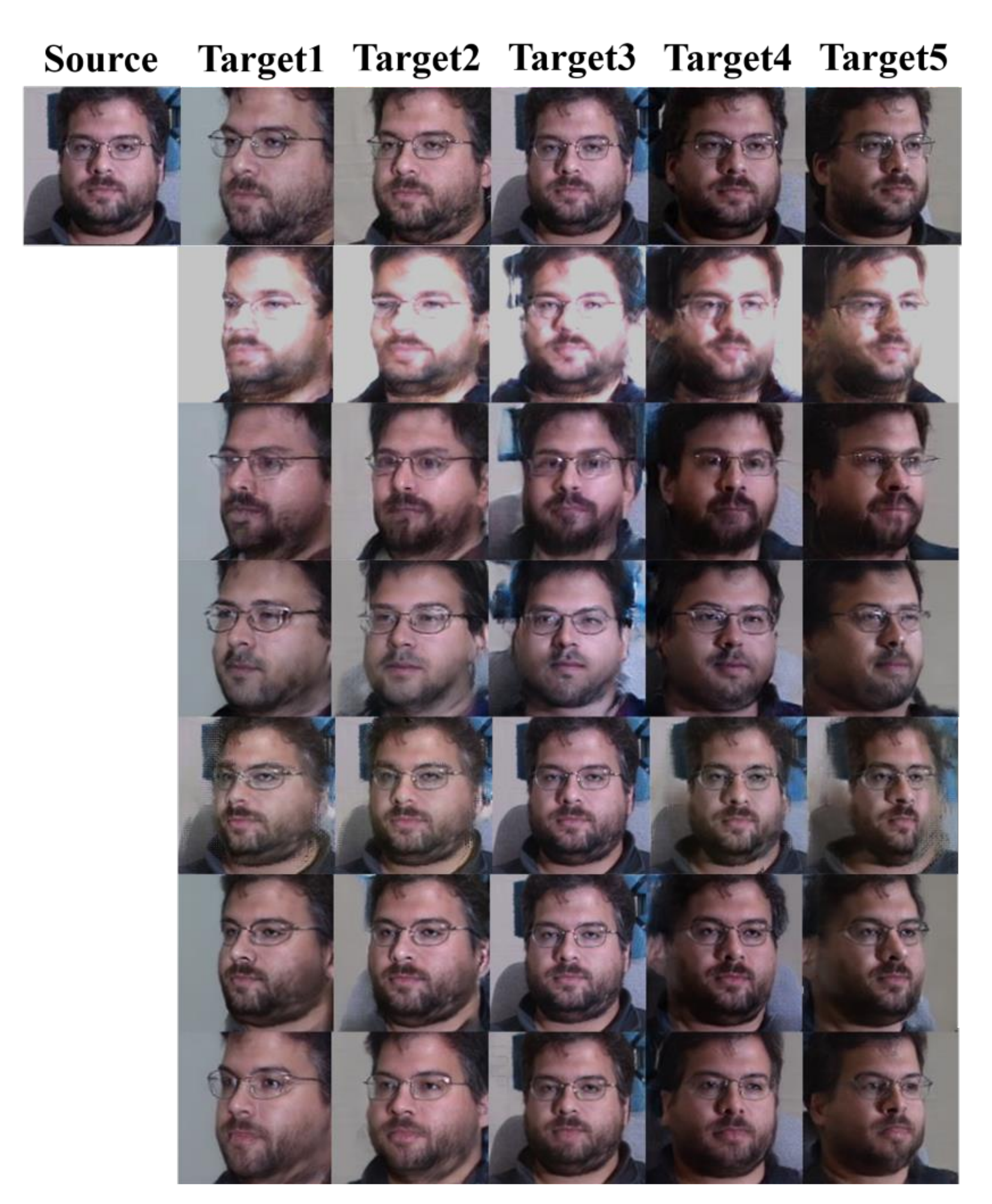}
		\end{minipage}
        \vspace{-0.2cm}
		\caption{Comparison on MultiPIE. For each image, the top row is the ground truth while the 2nd to 6th rows are generated by MV \cite{sun2018multi}, 
			CRGAN\cite{tian2018cr}, VIGAN\cite{xu2019view}, PONO\cite{li2019positional} and CDVAE\cite{yin2020novel} respectively. The last row is generated by our ID-Unet.
			}
		\label{fig:com_multiPIE}
        \vspace{-0.3cm}
	\end{figure*}
	\begin{figure}
		\begin{center}   
			\includegraphics[height=7.8cm]{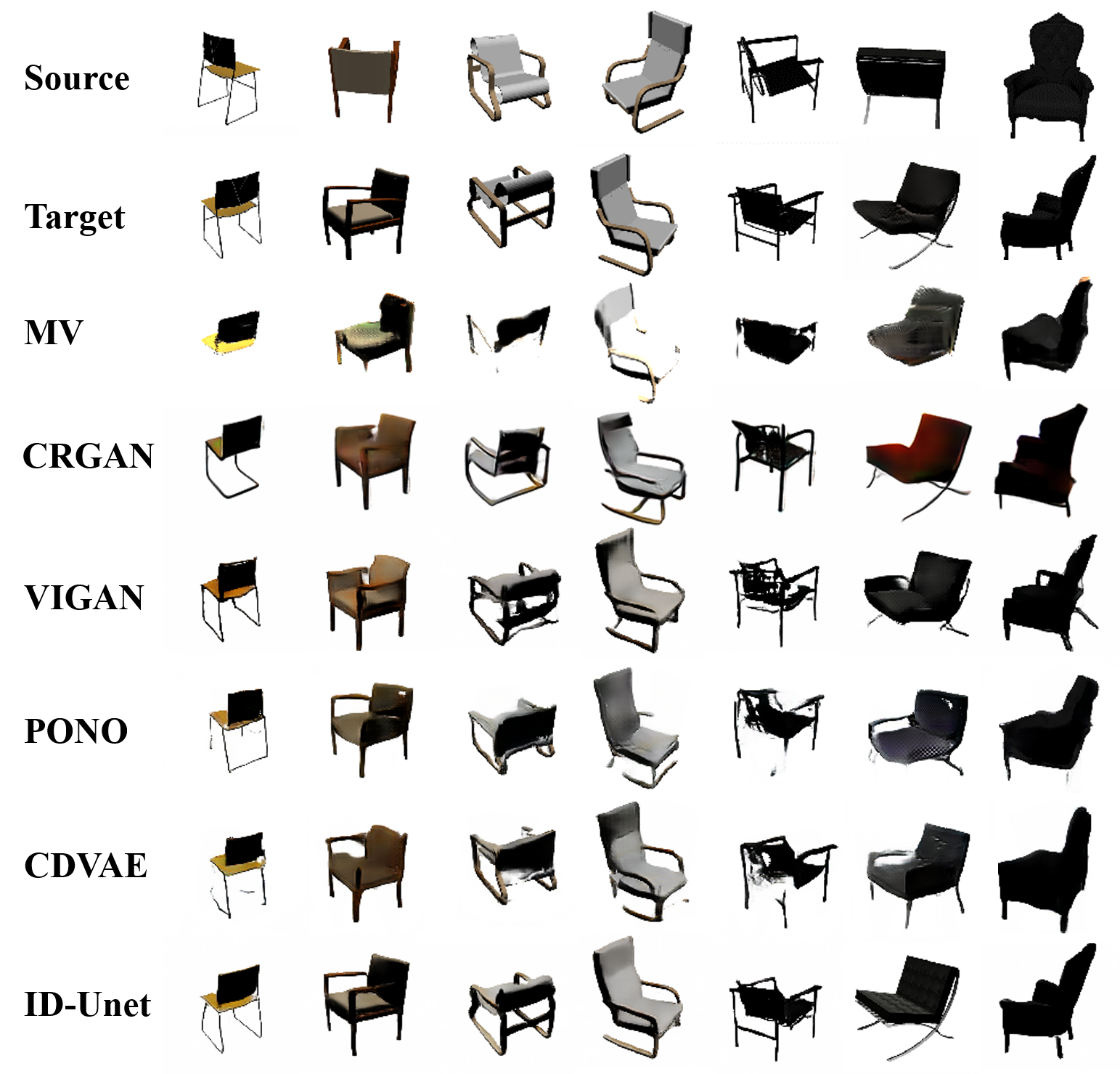} 
		\end{center}
        \vspace{-0.3cm}
		\caption{Comparison on 3D chair. The 1st and 2nd rows are the source and target images while the 3rd to 7th rows are generated by  MV \cite{sun2018multi}, 
		CRGAN \cite{tian2018cr}, VIGAN \cite{xu2019view}, PONO \cite{li2019positional} and CDVAE \cite{yin2020novel} respectively. The last row is generated by ID-Unet.}
		\label{fig:com_chair}
        \vspace{-0.3cm}
	\end{figure}
	
	\subsection{Visualizations}
	\textbf{Optical flow.} In Figure \ref{fig:flow_multiPIE}, the source image is translated into 3 target views. The 3rd row is the result from soft flow, which is converted into a 2-channel hard flow for visualization by taking out the most relevant coordinate. We find that the absolute value of the soft 
	flow is larger than the rest of the hard flow in the 1st, 2nd and 4th rows, which can be used to achieve overall deformation. The 4th row of KG flow in SCDM has the clear direction. 
	The magnitudes of residual hard flow in the 1st and 2nd row are smaller, showing that the feature progressively approaches the target view. 
	
	\textbf{Continuous view synthesis by interpolation.} To translate an image in an unseen view, we linearly interpolate the two conditions 
	to get an arbitrary angle image that does not exist in the dataset. Figure \ref{fig:interpolation} shows that our model is smooth enough to achieve view morphing. 
	
	\textbf{Visual comparisons with previous works.} As shown in Figure \ref{fig:com_multiPIE} and \ref{fig:com_chair}, ID-Unet 
	can accurately achieve the view synthesis while effectively maintain the source 
	contents, 
	\emph{e.g.}, the face ID 
	and the chair style. 
	The quantitative results in Table \ref{Table1} can also confirm the effectiveness.
	The 
	results from MV \cite{sun2018multi} are excessively bright, and it has problems such as ghosting for difficult samples. 
	VIGAN \cite{xu2019view} and CRGAN \cite{tian2018cr} have good results 
	on simple samples, but 
	they can not maintain the original structure for complex chairs, and synthesize the facial details like eyes 
	in the 2nd facial image. 
	PONO \cite{li2019positional} and CDVAE \cite{yin2020novel} have good ability to keep the source content, 
	but their models 
	do not understand the structure of complex objects. For example, the 3rd face in Figure \ref{fig:com_multiPIE} cannot achieve reasonable translation. 
	
	\section{Conclusion}
	This paper presents the ID-Unet to perform the view synthesis. It iteratively makes the deformation on the encoder features from different layers, and connects them into the decoder to complement the content details. To achieve the view translation, we design the SCDM and HCDM to align the feature from the source view to the target. Both the modules take the encoder and decoder features as well as the view condition vector as the inputs, compare the features to give either the soft or hard flow, and warp the encoder feature according to it. Since the flows are computed from features of different sizes, we accumulate them across resolutions and use the current flow to coarsely align the encoder feature first, and then estimate the residuals flow to refine it. Experiments show the effectiveness of the proposed model on two different datasets.

    \appendix
	
	\section{More Details on Network Architecture }
	 
\begin{figure}[h]
		\begin{center} 
			\includegraphics[width=0.46\textwidth]{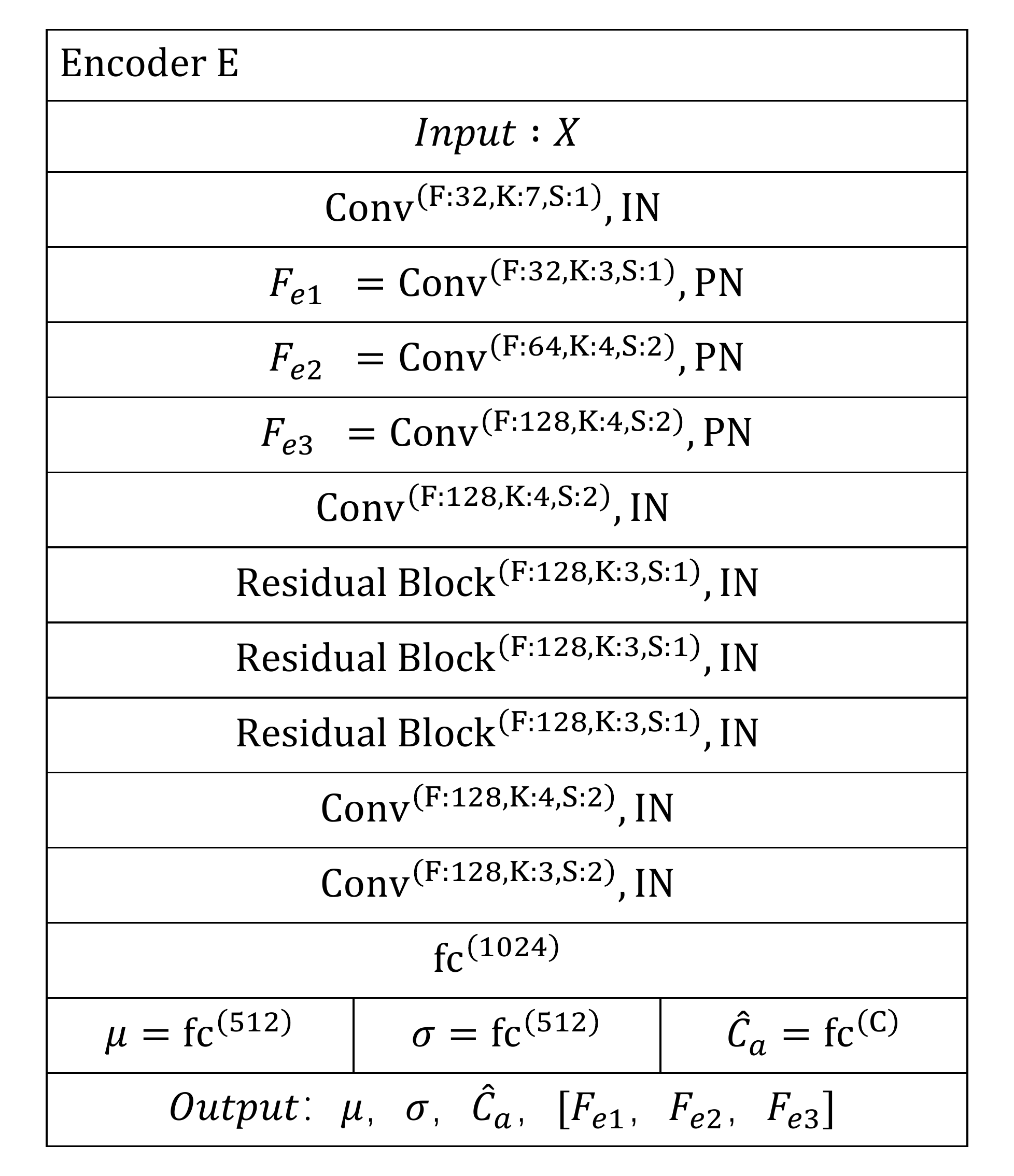}
		\end{center}
		\caption{The structure of Encoder. In $\mathrm{E}$, except the PN (positional normalization) \cite{li2019positional} used in the shallow layers ($F_{ei}, i=1,2,3$), the rest adopt the IN (instance normalization) \cite{ulyanov2016instance}.}
		\label{fig:encoder}
	\end{figure}

\begin{figure}[h]
		\begin{center} 
			\includegraphics[width=0.45\textwidth]{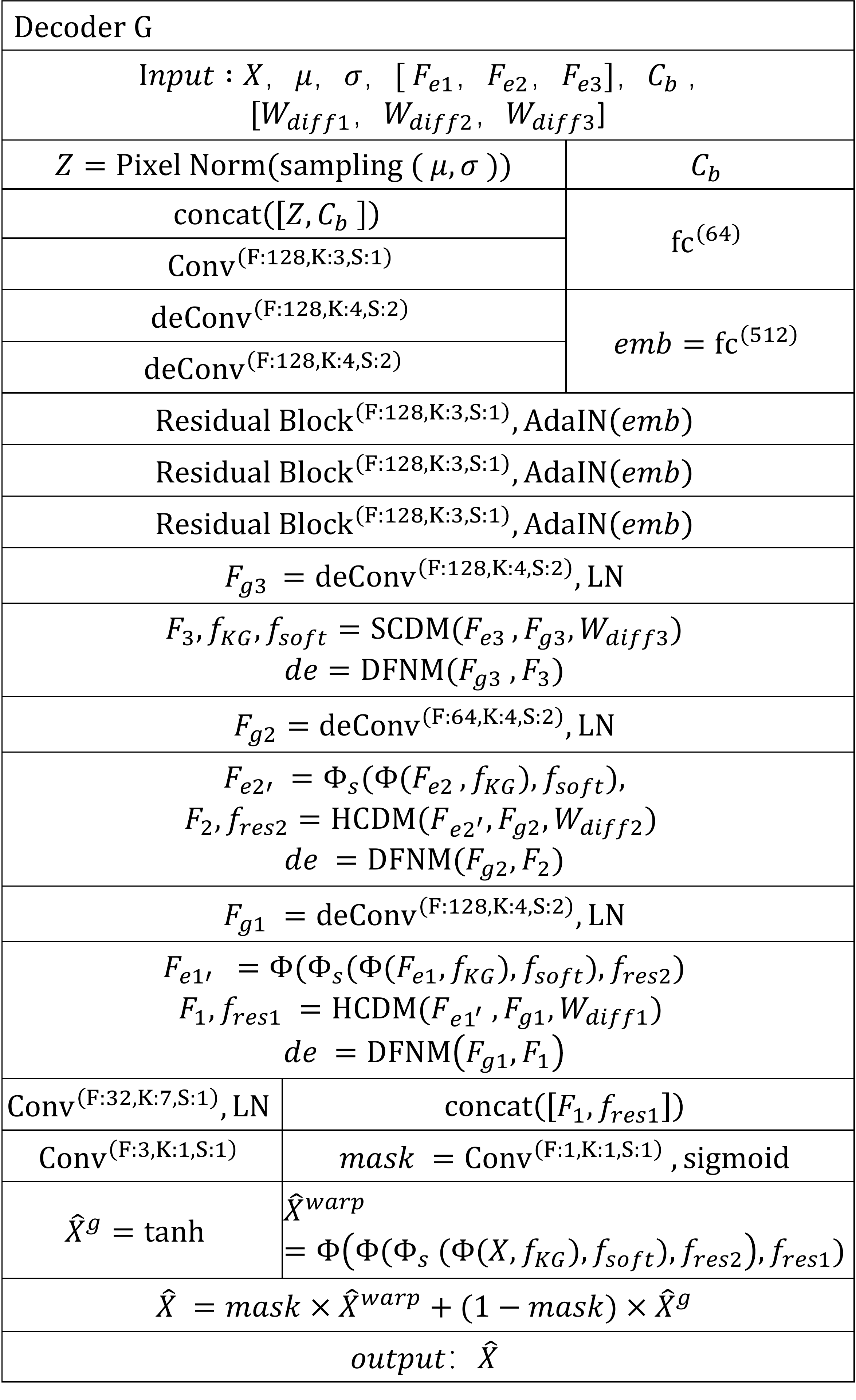}
		\end{center}
		\caption{The structure of Decoder. The final $\hat{X}$ is mixed by two parts. One is the $\hat{X}^{warp}$, obtained by the soft ($\Phi_s$) and hard ($\Phi$) deformation on the source $X$, and the other $\hat{X}^g$ is the output of the generator. Where the results of SCDM deformation ($F_{3}$) and HCDM  deformation ($F_{2}, F_{1}$) affects $F_{gi}, i=1,2,3$ in the form of DFNM \cite{yin2020novel}.}
		\label{fig:decoder}
	\end{figure}

\begin{figure}[t]
		\begin{center} 
			\includegraphics[width=0.48\textwidth]{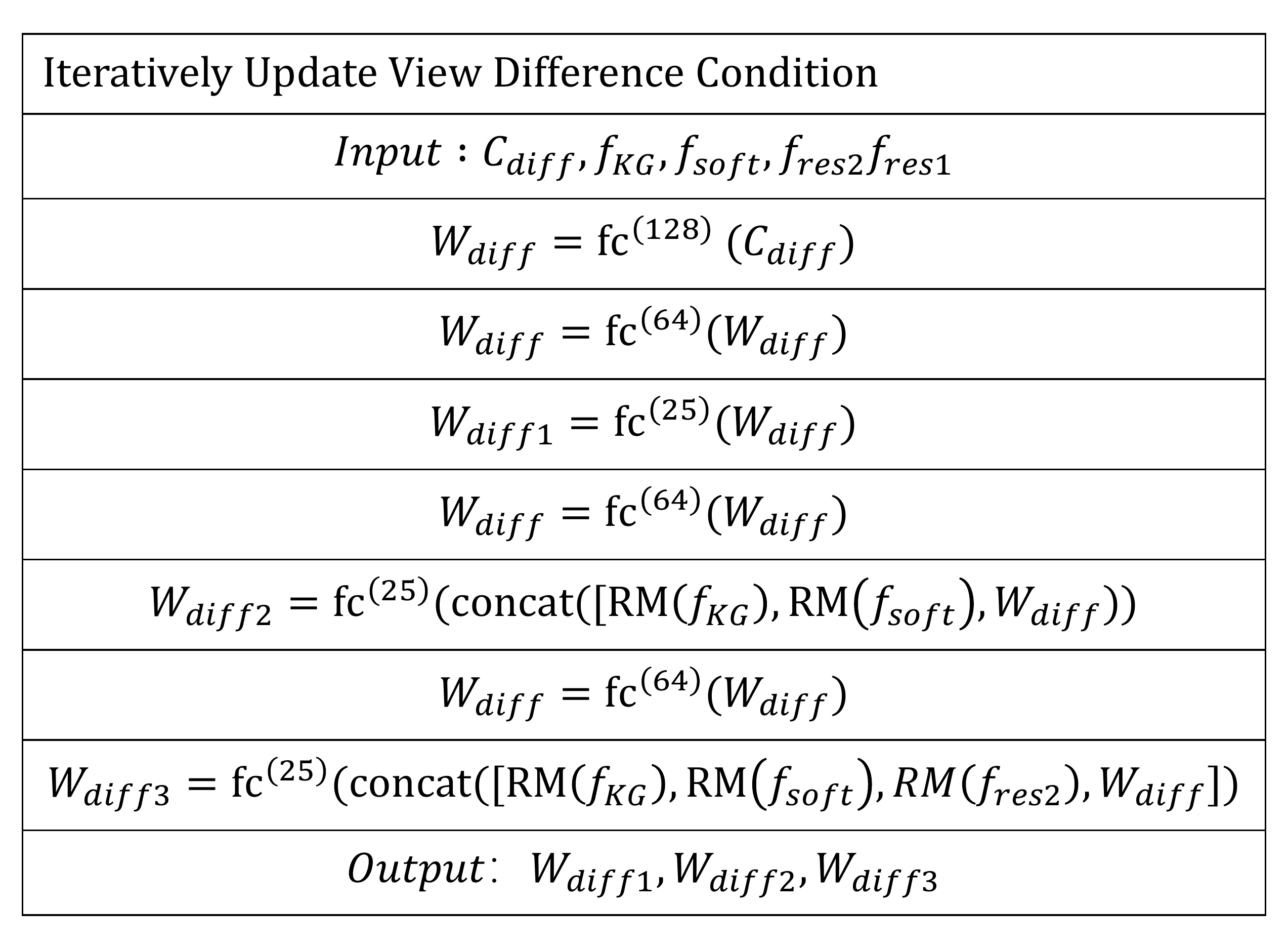}
		\end{center}
		\caption{The structure of Iterative conditional branch. Here RM is the reduce mean operation along the two spatial dimensions.}
		\label{fig:MLP}
	\end{figure}

\begin{figure}[t]
		\begin{center} 
			\includegraphics[width=0.48\textwidth]{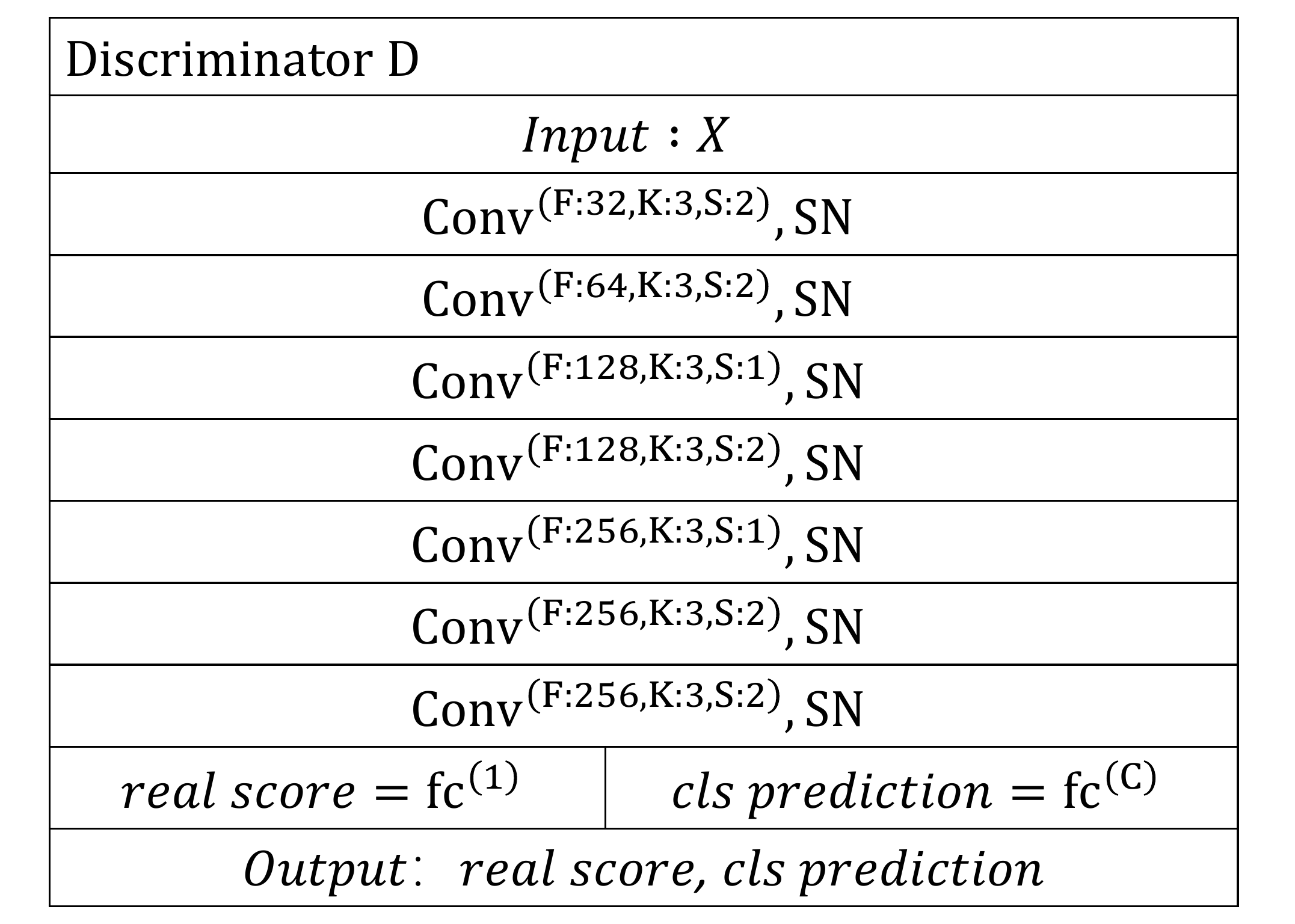}
		\end{center}
		\caption{The structure of Discriminator. In $\mathrm{D}$, 
		the SN (spectral normalization) \cite{miyato2018spectral} is applied to all layers.}
		\label{fig:discriminator}
	\end{figure}

	In this section, we give the specific details of network structure. Figure \ref{fig:encoder}, \ref{fig:decoder}, \ref{fig:MLP} and \ref{fig:discriminator} are the network structures of the encoder $\mathrm{E}$ 
	, the decoder $\mathrm{G}$, the iterative view difference condition branch and the discriminator $\mathrm{D}$, respectively. In Conv and Residual block, F, K and S respectively represent the number of kernels, the size of the convolution kernel and the stride.
	We use the ADAM \cite{kingma2014adam} with learning rates 0.0002 and set $\beta_1$=0, $\beta_2$=0.9. We will release our code if this paper is accepted.


	\section{More Visualized Results}
	\subsection{
	The 
	results of view translation}
\begin{figure*}[b]
		\begin{center} 
			\includegraphics[width=0.95\textwidth]{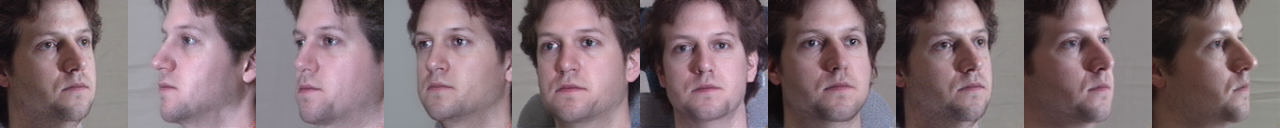}
			\includegraphics[width=0.95\textwidth]{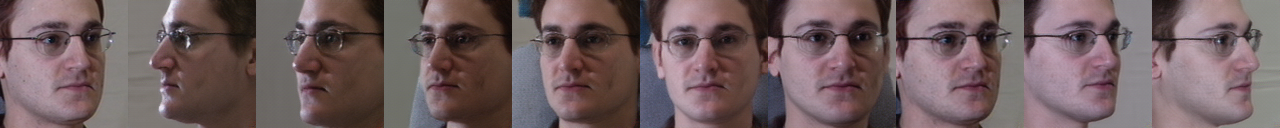}
			\includegraphics[width=0.95\textwidth]{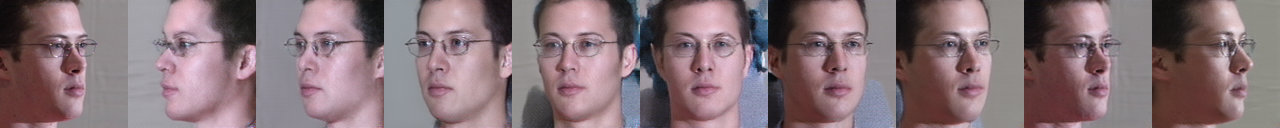}
			\includegraphics[width=0.95\textwidth]{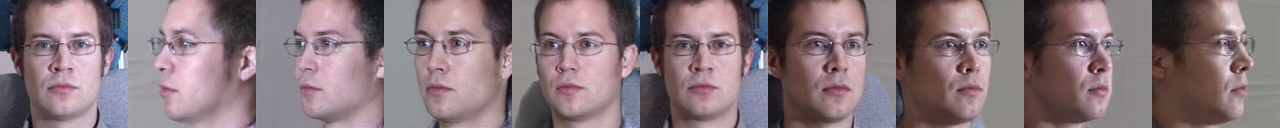}
		    \includegraphics[width=0.95\textwidth]{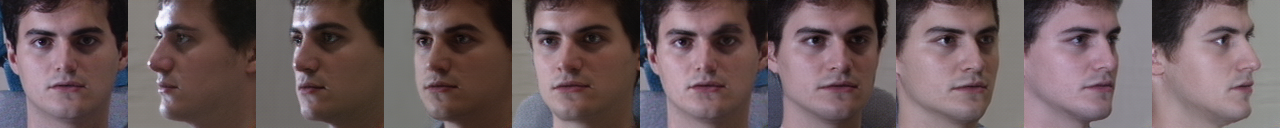}
			\includegraphics[width=0.95\textwidth]{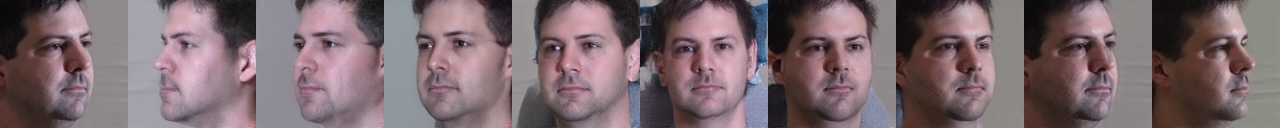}
			\includegraphics[width=0.95\textwidth]{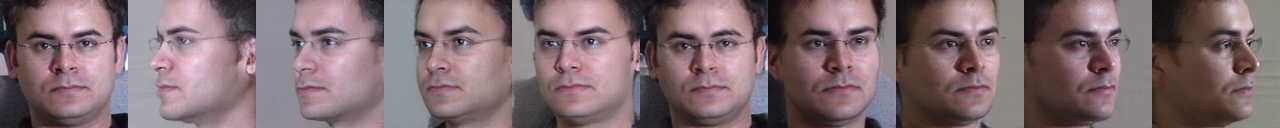}
			\includegraphics[width=0.95\textwidth]{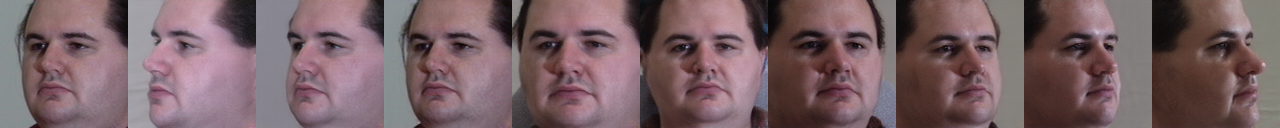}
		    \includegraphics[width=0.95\textwidth]{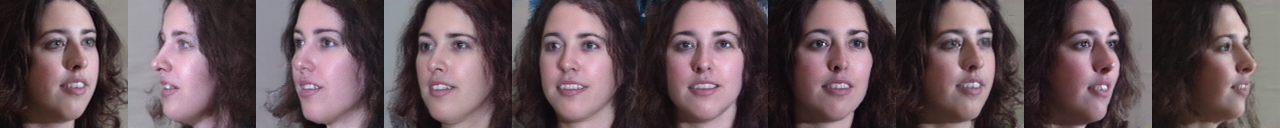}
			\includegraphics[width=0.95\textwidth]{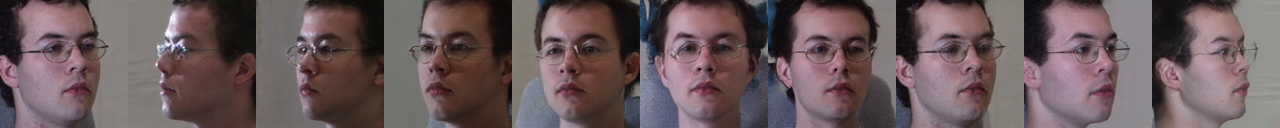}
			\includegraphics[width=0.95\textwidth]{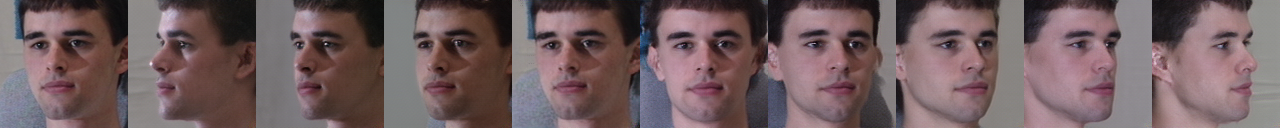}
		\end{center}
		\caption{More results on MultiPIE dataset\cite{gross2010multi}. The 1st column is the source image, and the remaining columns are the generated images under different target views.}
		\label{fig:multiPIE1}
	\end{figure*}

\begin{figure*}
		\begin{center} 
			\includegraphics[width=0.95\textwidth]{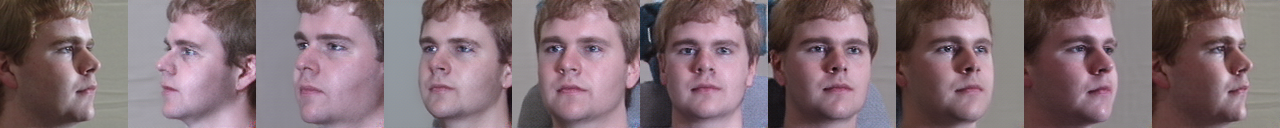}
			\includegraphics[width=0.95\textwidth]{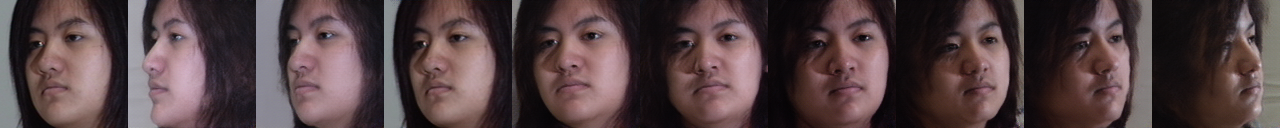}
			\includegraphics[width=0.95\textwidth]{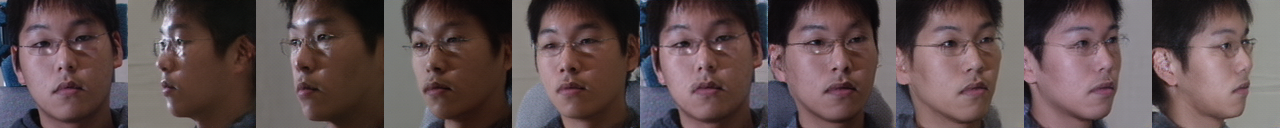}
			\includegraphics[width=0.95\textwidth]{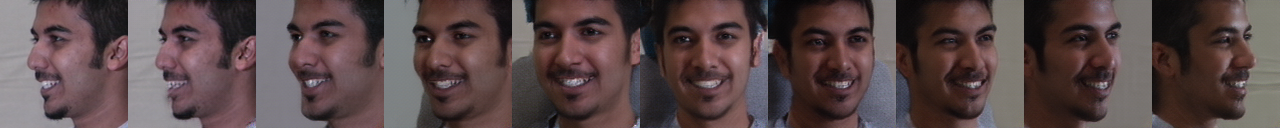}
			\includegraphics[width=0.95\textwidth]{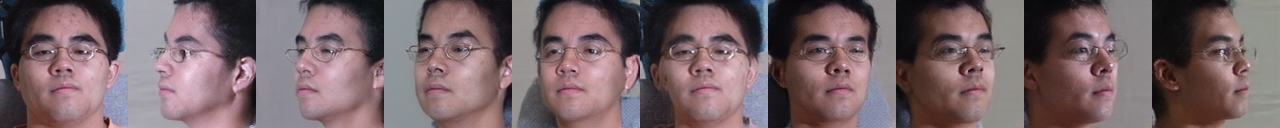}
			\includegraphics[width=0.95\textwidth]{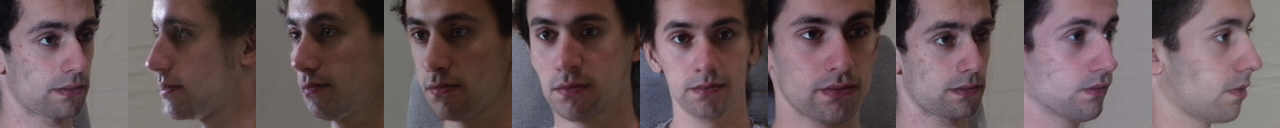}
			\includegraphics[width=0.95\textwidth]{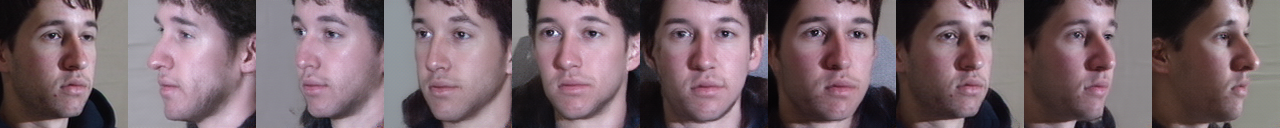}
			\includegraphics[width=0.95\textwidth]{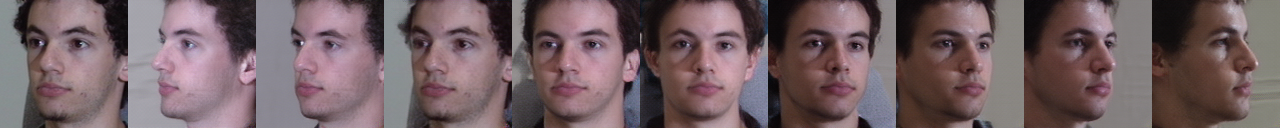}
			\includegraphics[width=0.95\textwidth]{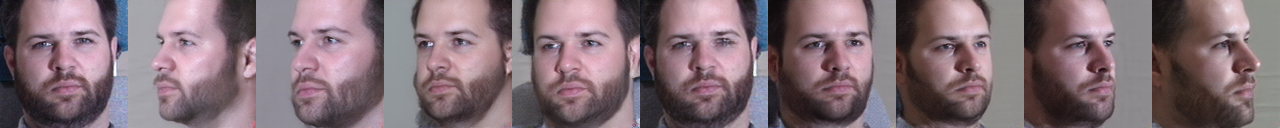}
			\includegraphics[width=0.95\textwidth]{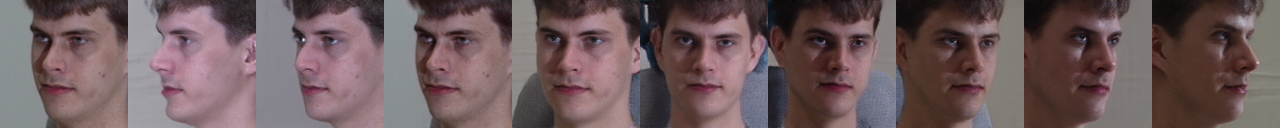}
			\includegraphics[width=0.95\textwidth]{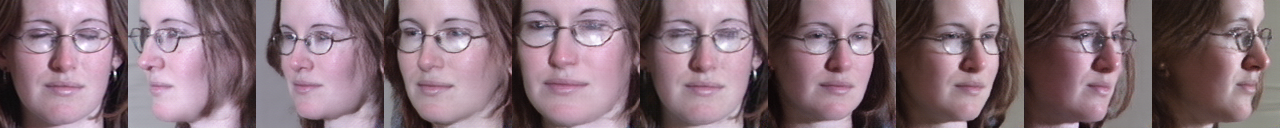}
		\end{center}
		\caption{More results on MultiPIE dataset\cite{gross2010multi}. The 1st column is the source image, and the remaining columns are the generated images under different target views.}
		\label{fig:multiPIE2}
	\end{figure*}

\begin{figure*}
		\begin{center} 
			\includegraphics[width=0.9\textwidth]{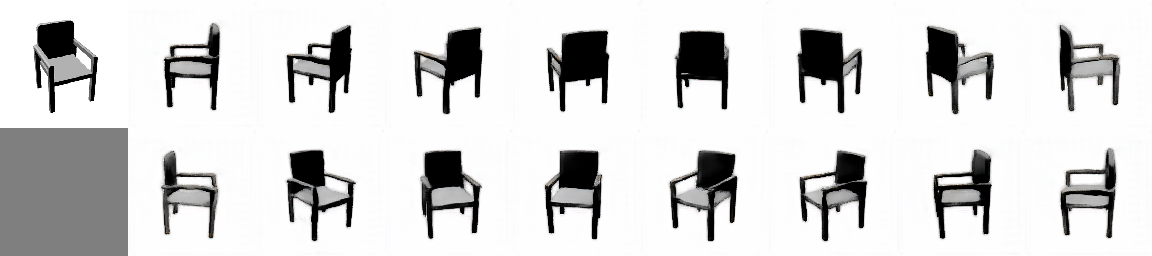}
			\includegraphics[width=0.9\textwidth]{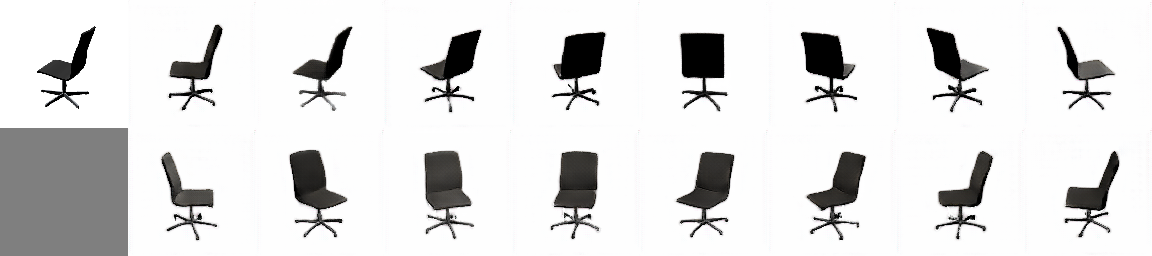}
			\includegraphics[width=0.9\textwidth]{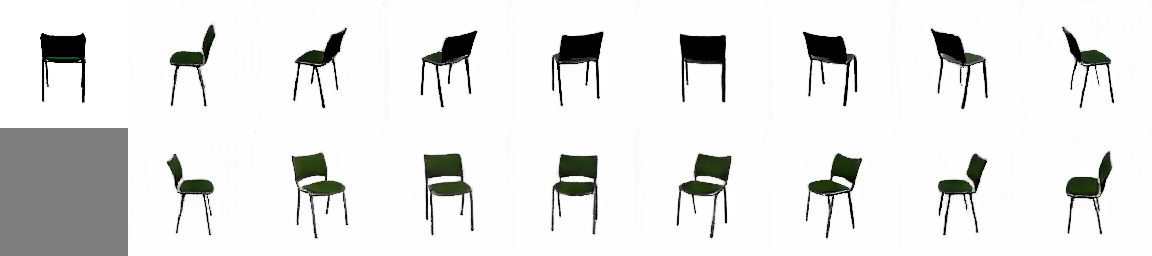}
			\includegraphics[width=0.9\textwidth]{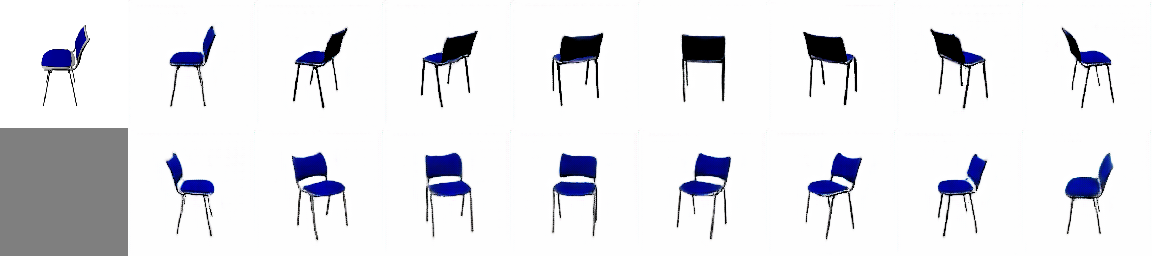}
			\includegraphics[width=0.9\textwidth]{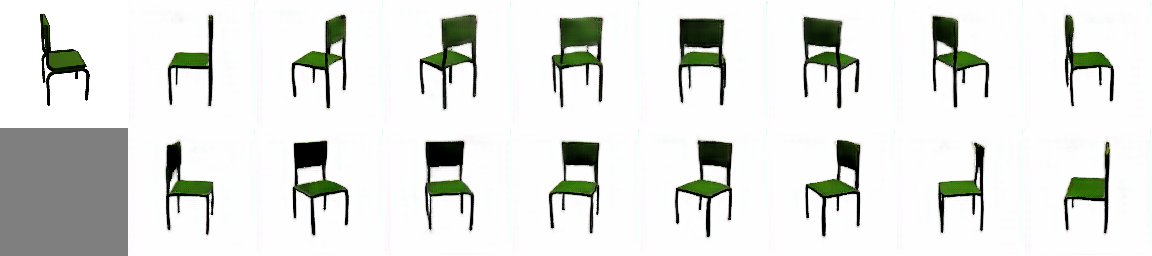}
			\includegraphics[width=0.9\textwidth]{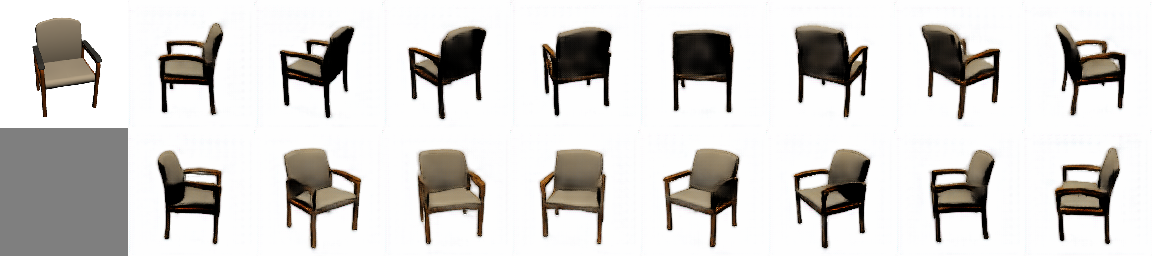}
		\end{center}
		\caption{More results on 3D chair\cite{aubry2014seeing} dataset. The 1st column is the source image, and the remaining columns are the generated images under different target views.}
		\label{fig:3d chair1}
	\end{figure*}

\begin{figure*}
		\begin{center} 
			\includegraphics[width=0.9\textwidth]{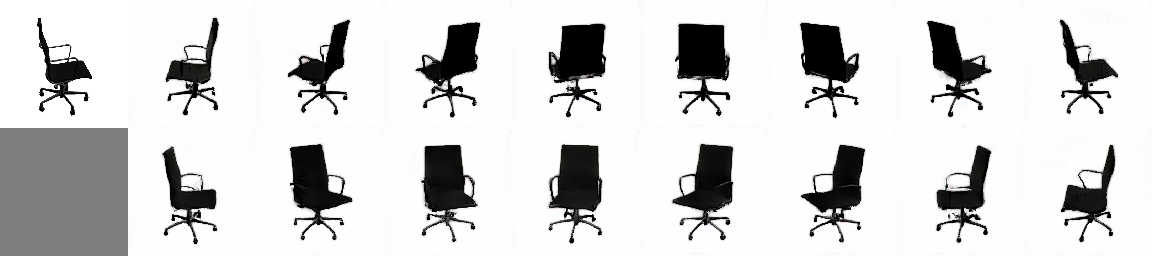}
			\includegraphics[width=0.9\textwidth]{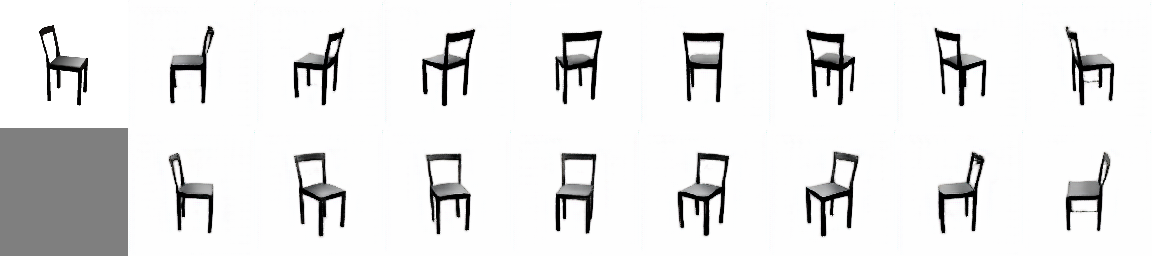}
			\includegraphics[width=0.9\textwidth]{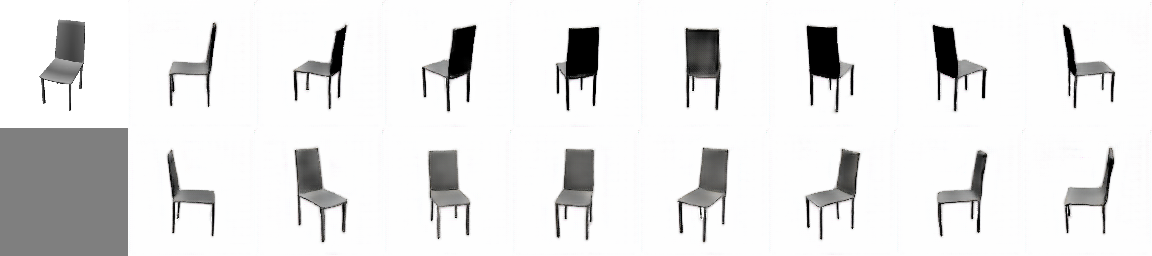}
			\includegraphics[width=0.9\textwidth]{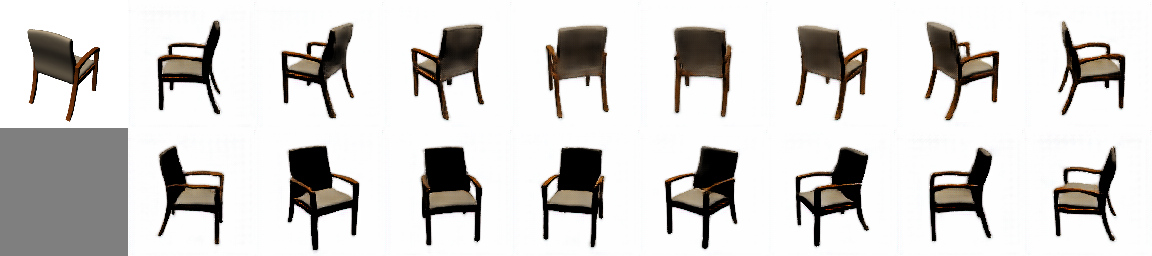}
			\includegraphics[width=0.9\textwidth]{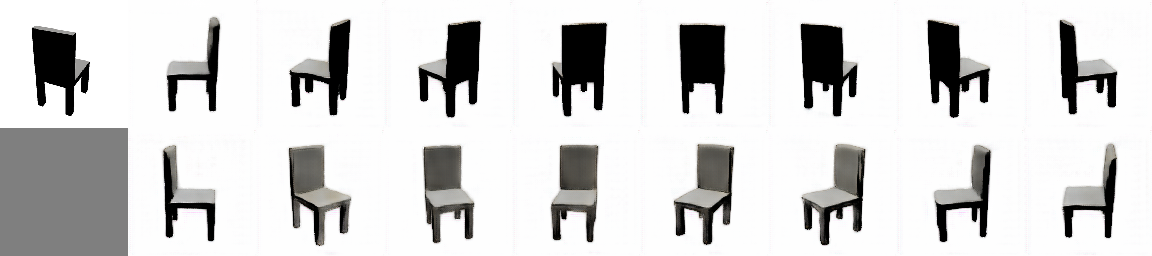}
			\includegraphics[width=0.9\textwidth]{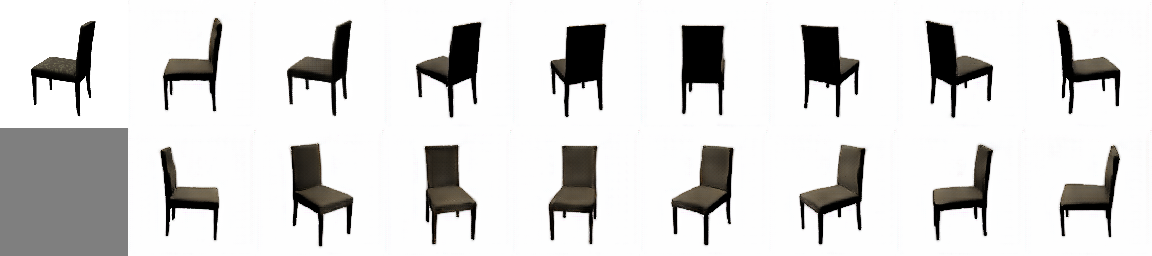}
		\end{center}
		\caption{More results on 3D chair\cite{aubry2014seeing} dataset. The 1st column is the source image, and the remaining columns are the generated images under different target views.}
		\label{fig:3d chair2}
	\end{figure*}

\begin{figure*}
		\begin{center} 
			\includegraphics[width=0.9\textwidth]{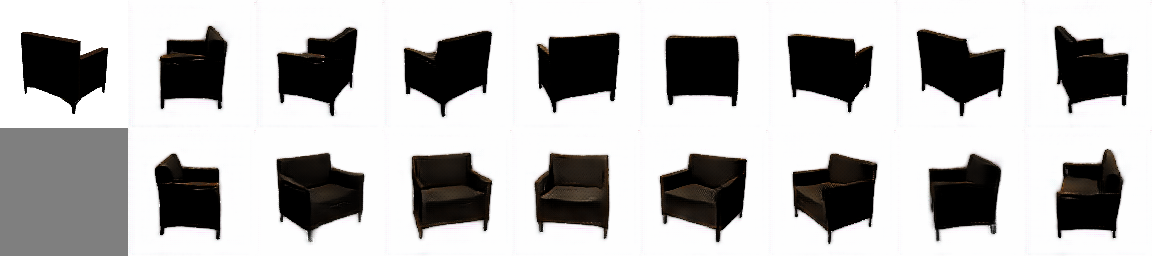}
			\includegraphics[width=0.9\textwidth]{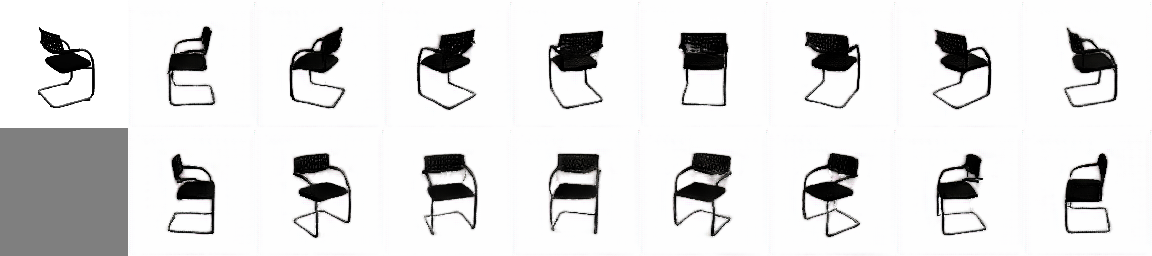}
			\includegraphics[width=0.9\textwidth]{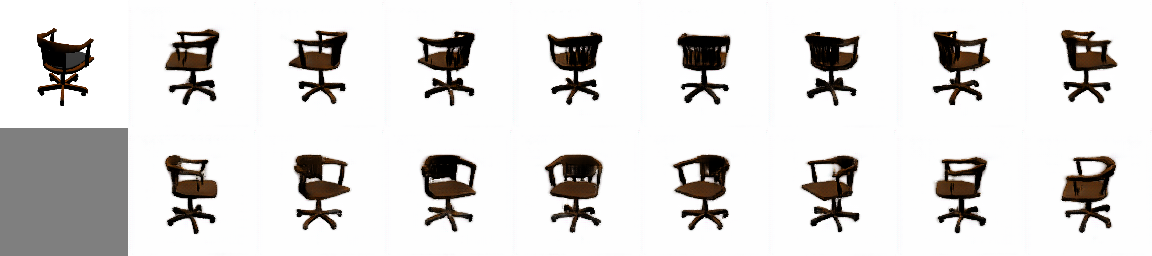}
			\includegraphics[width=0.9\textwidth]{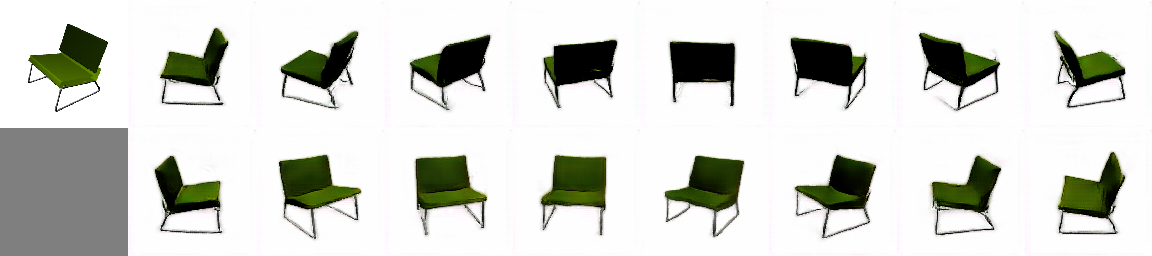}
			\includegraphics[width=0.9\textwidth]{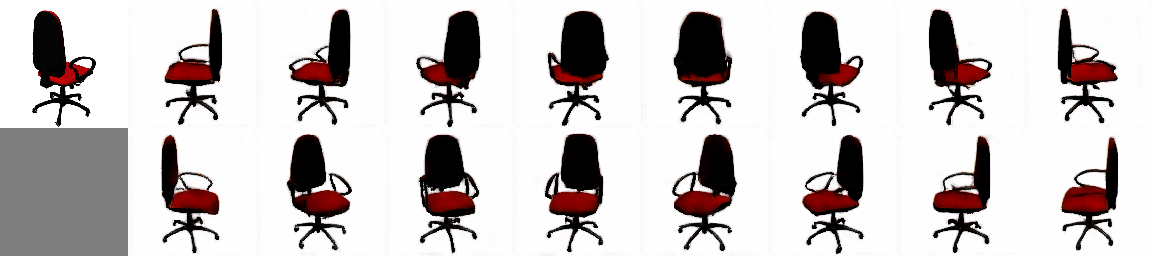}
			\includegraphics[width=0.9\textwidth]{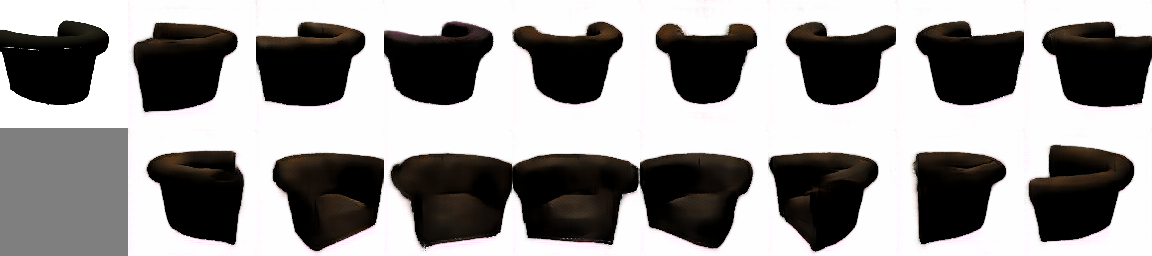}
		\end{center}
		\caption{More results on 3D chair\cite{aubry2014seeing} dataset. The 1st column is the source image, and the remaining columns are the generated images under different target views.}
		\label{fig:3d chair3}
	\end{figure*}

\begin{figure*}
		\begin{center} 
			\includegraphics[width=0.9\textwidth]{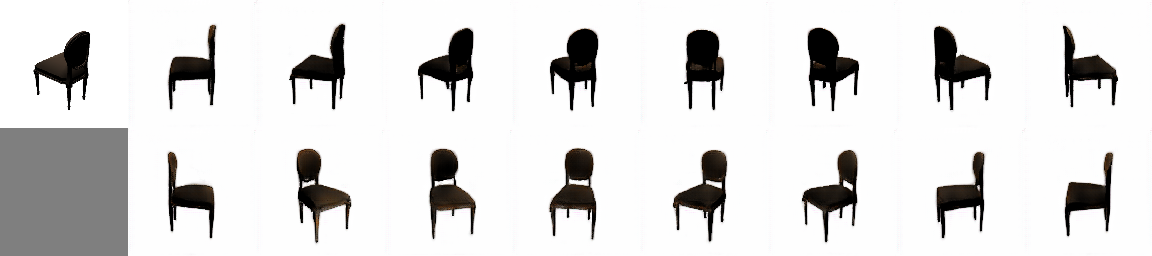}
			\includegraphics[width=0.9\textwidth]{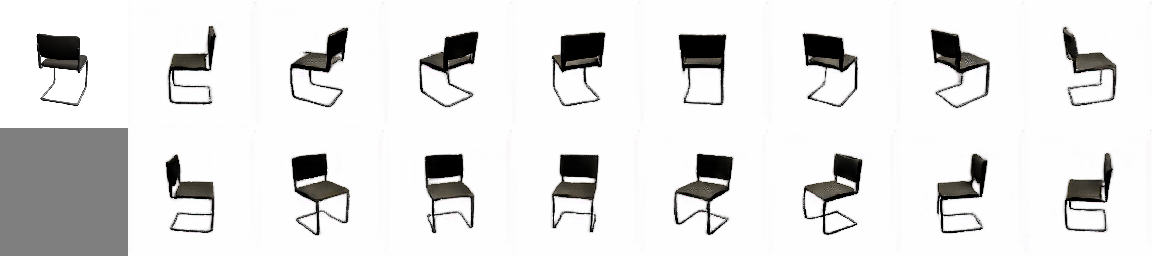}
			\includegraphics[width=0.9\textwidth]{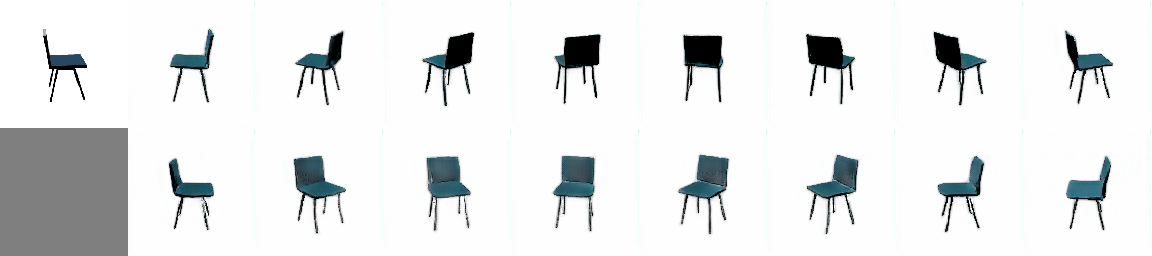}
			\includegraphics[width=0.9\textwidth]{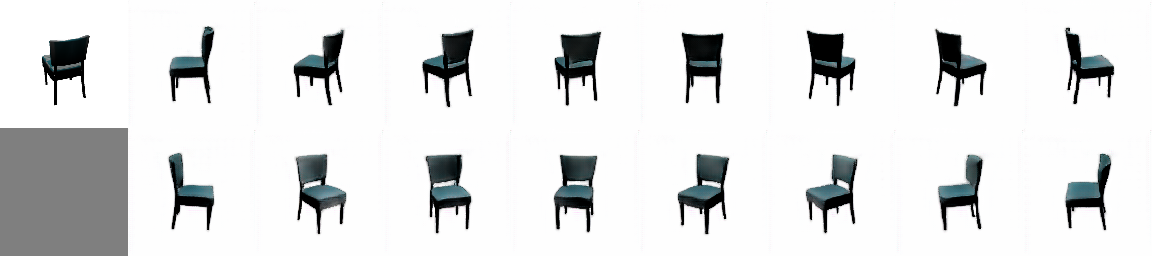}
			\includegraphics[width=0.9\textwidth]{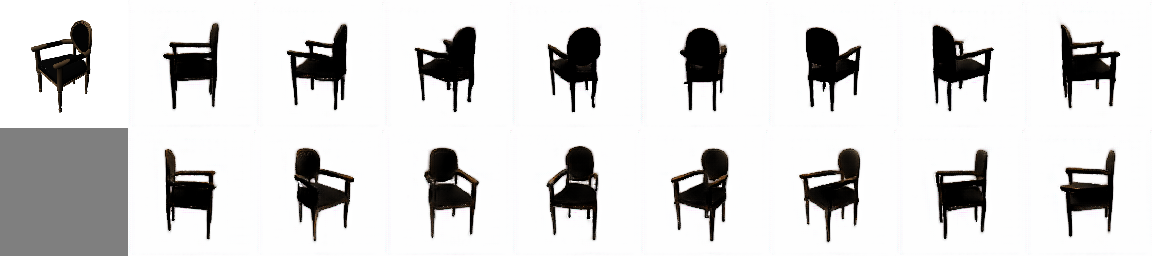}
			\includegraphics[width=0.9\textwidth]{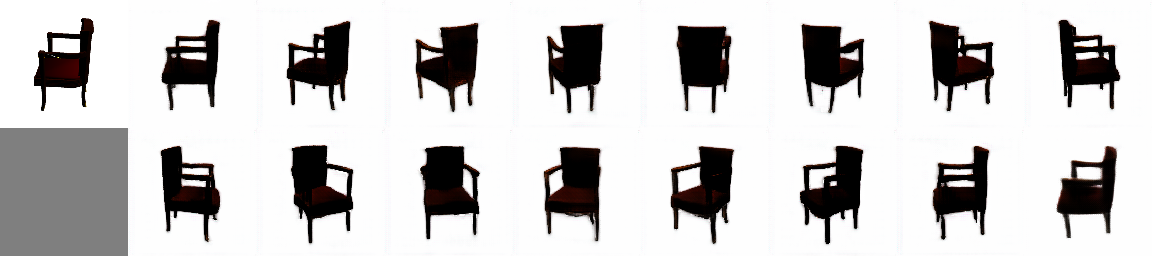}
		\end{center}
		\caption{More results on 3D chair\cite{aubry2014seeing} dataset. The 1st column is the source image, and the remaining columns are the generated images under different target views.}
		\label{fig:3d chair4}
	\end{figure*}

\begin{figure*}
		\begin{center} 
			\includegraphics[width=0.9\textwidth]{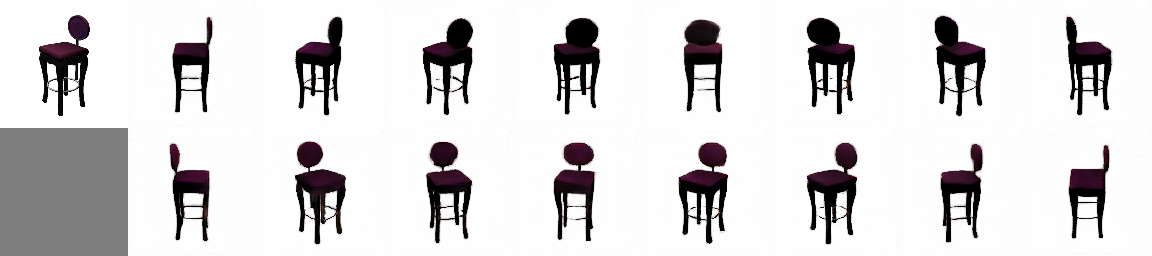}
			\includegraphics[width=0.9\textwidth]{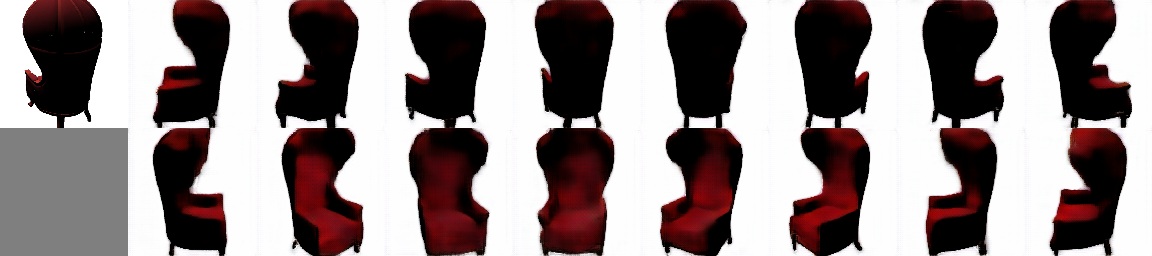}
			\includegraphics[width=0.9\textwidth]{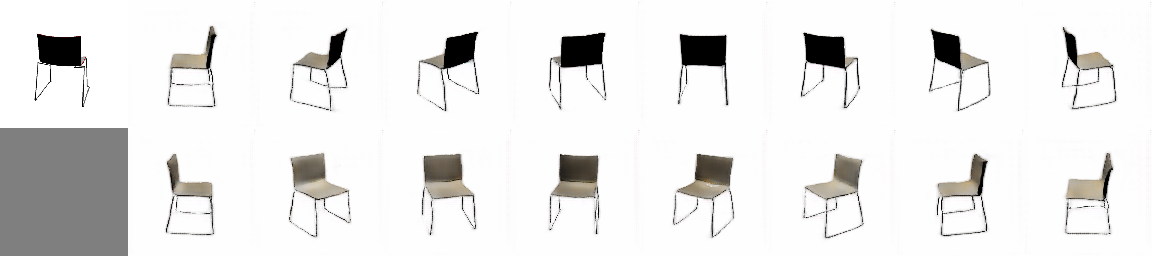}
			\includegraphics[width=0.9\textwidth]{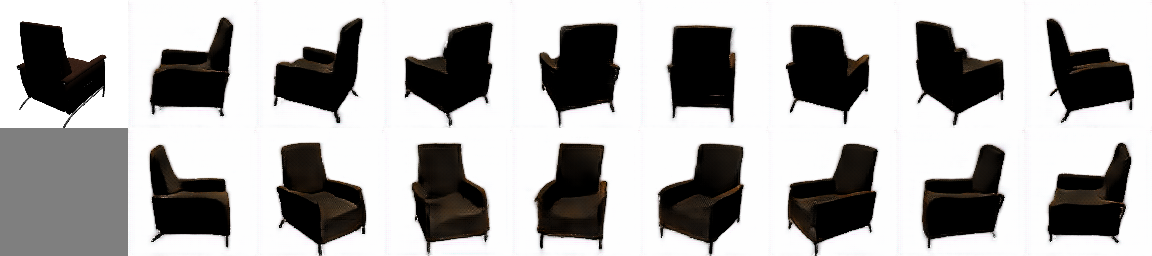}
			\includegraphics[width=0.9\textwidth]{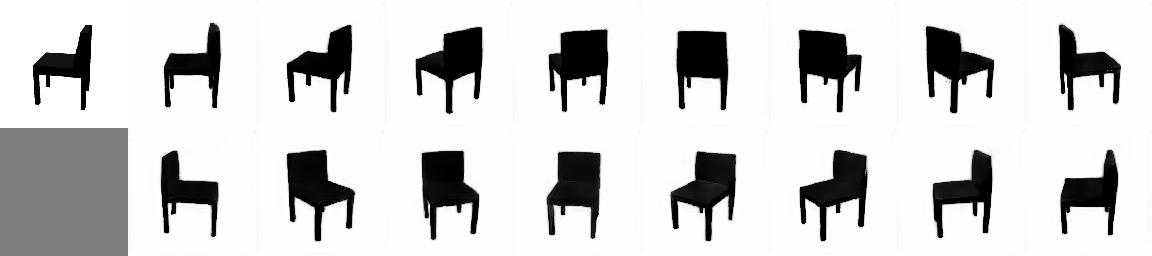}
			\includegraphics[width=0.9\textwidth]{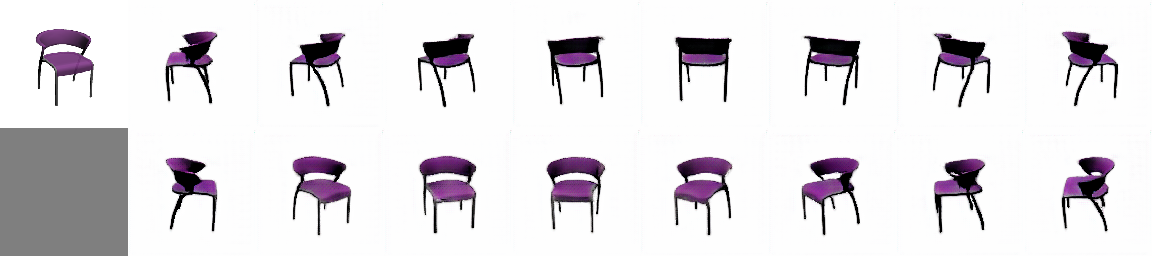}
		\end{center}
		\caption{More results on 3D chair\cite{aubry2014seeing} dataset. The 1st column is the source image, and the remaining columns are the generated images under different target views.}
		\label{fig:3d chair5}
	\end{figure*}

\begin{figure*}
		\begin{center} 
			\includegraphics[width=0.9\textwidth]{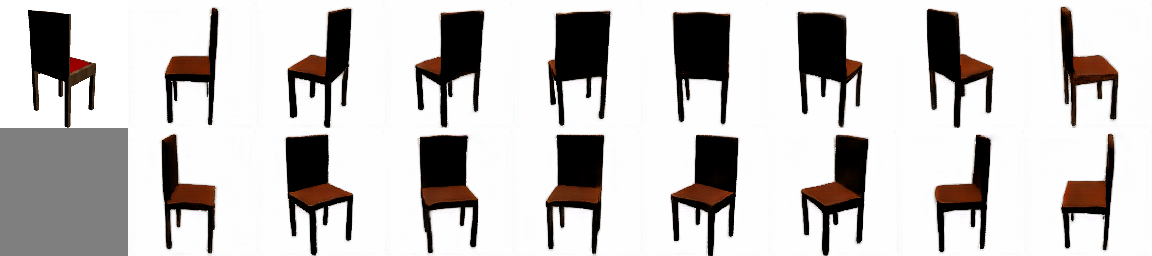}
			\includegraphics[width=0.9\textwidth]{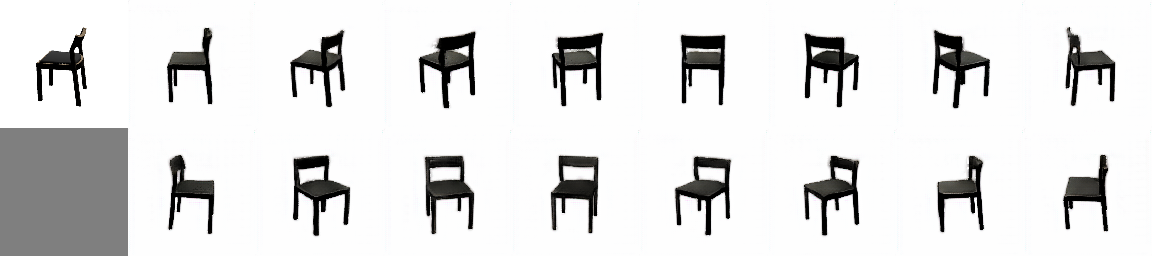}
			\includegraphics[width=0.9\textwidth]{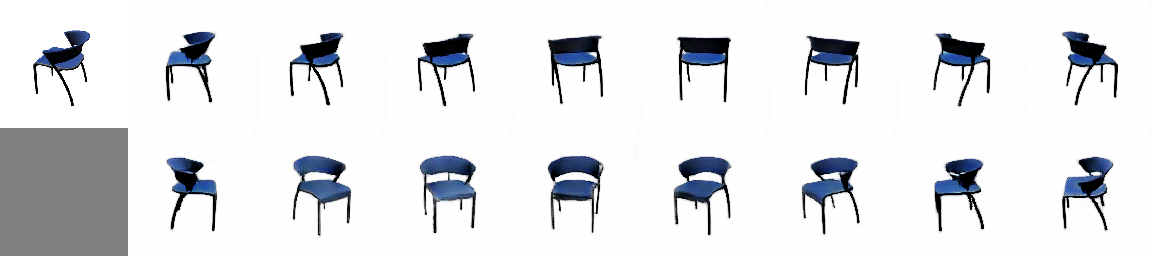}
			\includegraphics[width=0.9\textwidth]{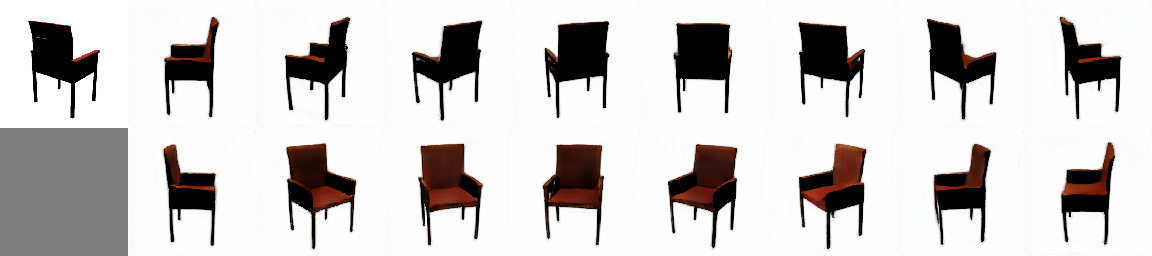}
			\includegraphics[width=0.9\textwidth]{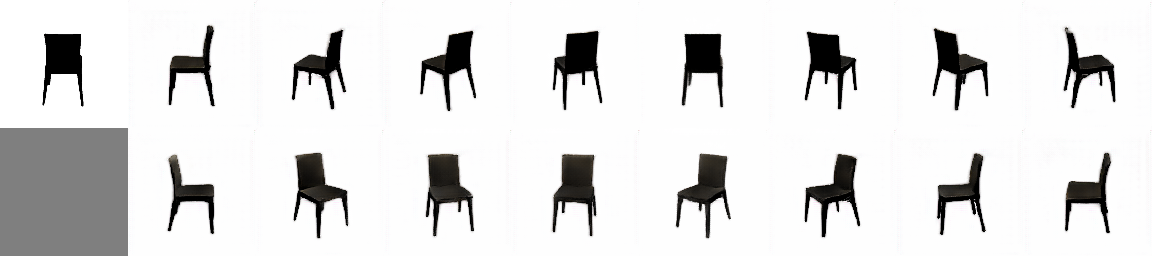}
			\includegraphics[width=0.9\textwidth]{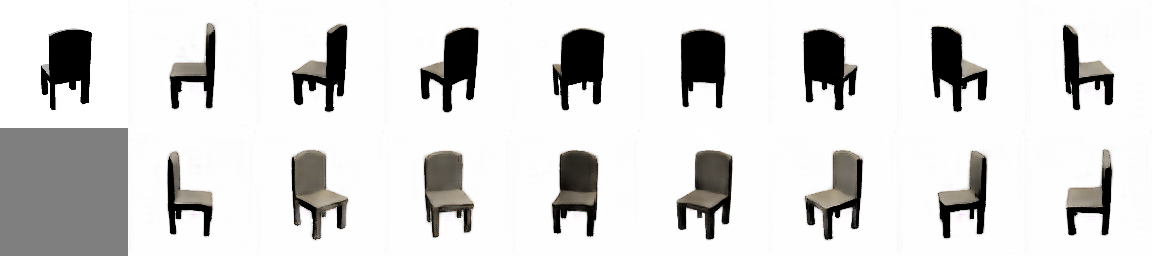}
		\end{center}
		\caption{More results on 3D chair\cite{aubry2014seeing} dataset. The 1st column is the source image, and the remaining columns are the generated images under different target views.}
		\label{fig:3d chair6}
	\end{figure*}

	Plenty of results of our method on the MultiPIE \cite{gross2010multi} dataset are shown in Figures \ref{fig:multiPIE1} and \ref{fig:multiPIE2}.
	Extra results on the 3D chair \cite{gross2010multi} dataset are shown in Figures \ref{fig:3d chair1}, \ref{fig:3d chair2}, \ref{fig:3d chair3}, \ref{fig:3d chair4}, \ref{fig:3d chair5} and \ref{fig:3d chair6}. Note that for all Figures, the 1st column is the source image, and the remaining columns are the generated images under different target views.
	
	\subsection{
	The visualization for the flow}
\begin{figure*}
	    \flushright
		\includegraphics[width=0.05\textwidth]{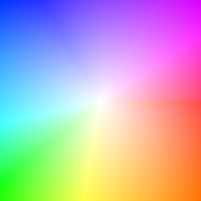}
		\begin{center} 
			\includegraphics[width=0.95\textwidth]{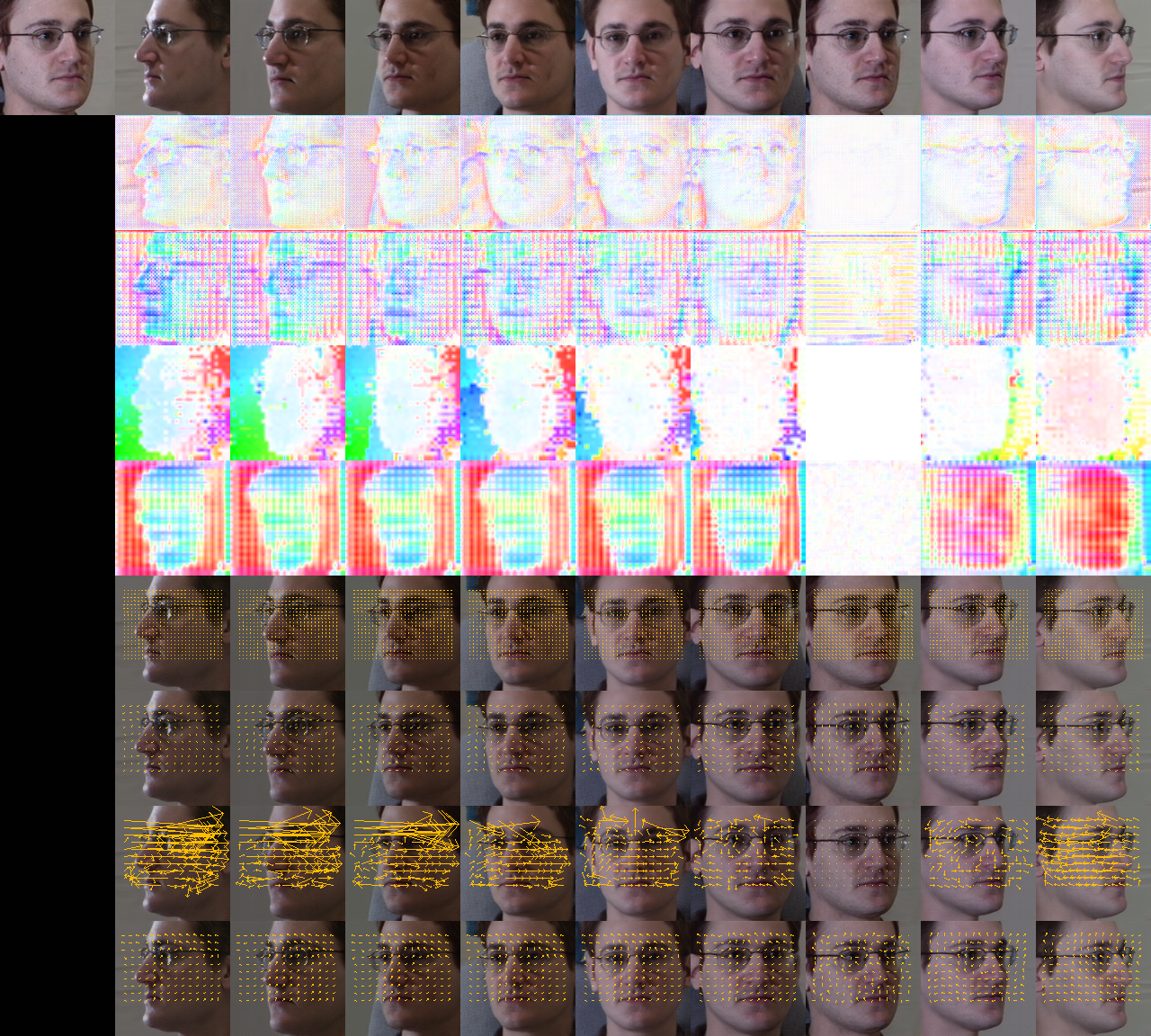}
		\end{center}
		\caption{The 1st row is the generated image from the source image (1st column) to various target views, the 2nd and 6th rows are Res hard flow (H×W), the 3rd and 7th rows are Res hard flow (H/2×W/2), the 4th and 8th  rows are Soft flow (H/4×W/4), and the 5th and 9th rows are KG flow (H/4×W/4).}
		\label{fig:flow1}
	\end{figure*}

\begin{figure*}
	    \flushright
		\includegraphics[width=0.05\textwidth]{flow_key.png}
		\begin{center}
			\includegraphics[width=0.95\textwidth]{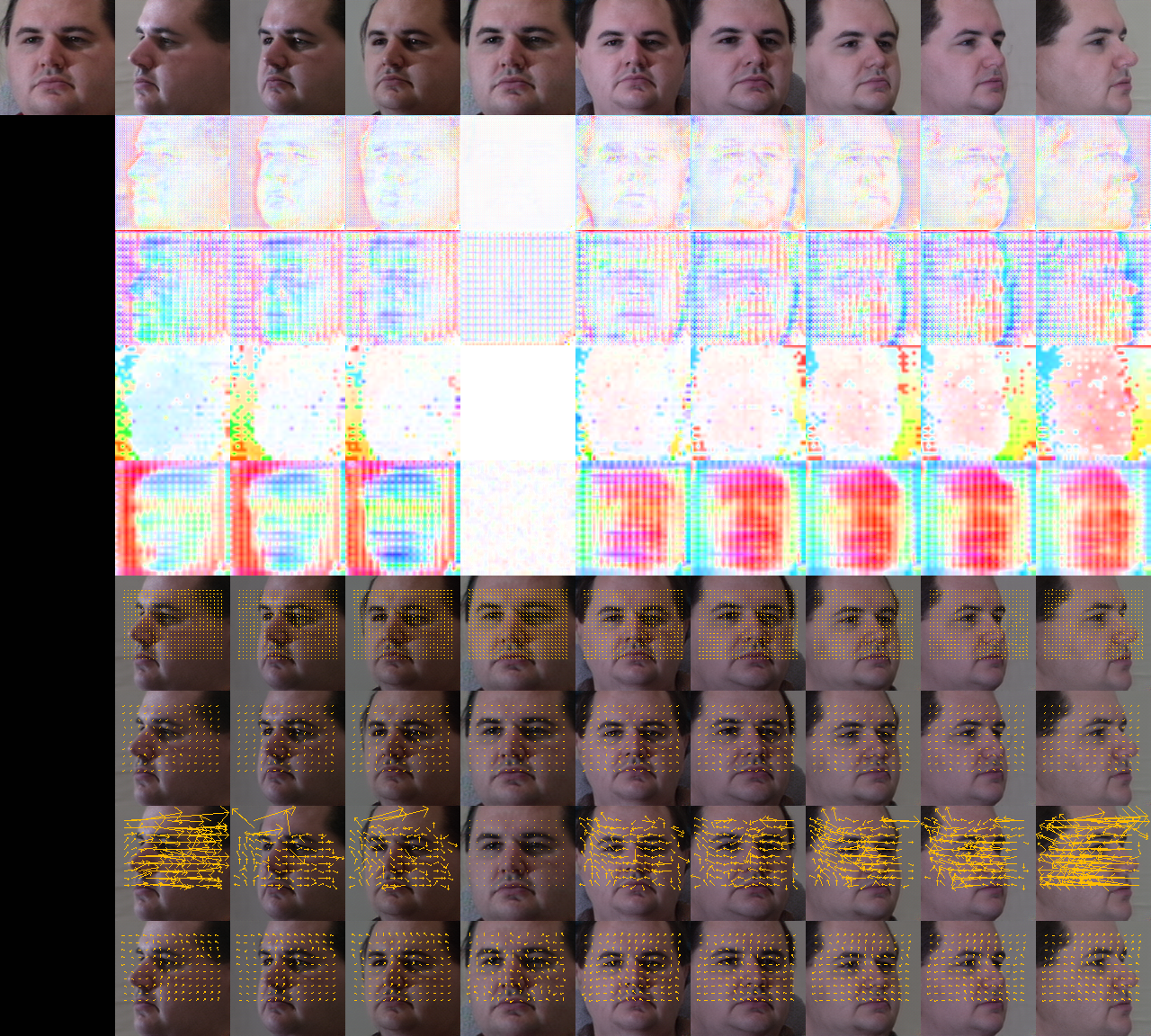} 
		\end{center}
		\caption{The 1st row is the generated image from the source image (1st column) to various target views, the 2nd and 6th rows are Res hard flow (H×W), the 3rd and 7th rows are Res hard flow (H/2×W/2), the 4th and 8th  rows are Soft flow (H/4×W/4), and the 5th and 9th rows are KG flow (H/4×W/4).}
		\label{fig:flow2}
	\end{figure*}

\begin{figure*}
	    \flushright
		\includegraphics[width=0.05\textwidth]{flow_key.png}
		\begin{center}
			\includegraphics[width=0.95\textwidth]{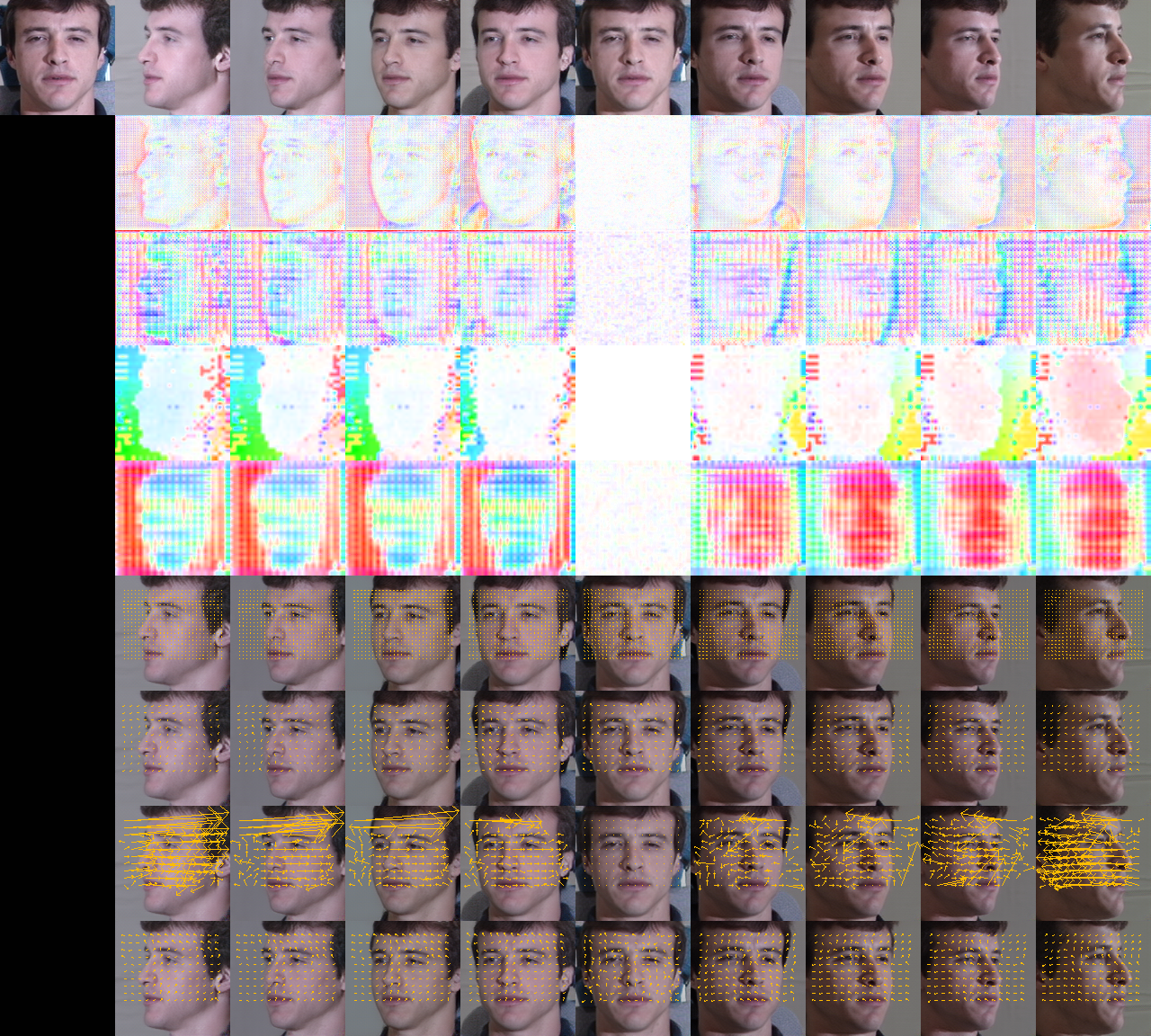}
		\end{center}
		\caption{The 1st row is the generated image from the source image (1st column) to various target views, the 2nd and 6th rows are Res hard flow (H×W), the 3rd and 7th rows are Res hard flow (H/2×W/2), the 4th and 8th  rows are Soft flow (H/4×W/4), and the 5th and 9th rows are KG flow (H/4×W/4).}
		\label{fig:flow3}
	\end{figure*}

	As shown in Figure \ref{fig:flow1}, \ref{fig:flow2} and \ref{fig:flow3}, the source image (1st column) is translated into 9 target views (2nd to 10th columns). We visualize the optical flows from various target views, namely Res hard flow, Soft flow and KG flow.\\
	
	\noindent\textbf{Res hard flow}\\
	The 2nd and 6th rows are the results 
	in full resolution. The 3rd and 7th rows are 
	in the half resolution. Both of them have more details, and the 
	flow amplitude is small. This indicates that they are used for the refinement and 
	supplement local details. Through observations, it can be found that the directions 
	for pixels are not exactly the same. 
	However, 
	most of them 
	are still consistent with the overall rotation. \emph{E.g.} the face region becomes light blue when it turns to the left, and 
	light red when turning to the right.\\
	
	\noindent\textbf{Soft flow}\\
	The Soft flow (in the quarter resolution) is shown in the 4th and 8th rows. It (the 8th row) has a large amplitude, which can better realize the whole rough view deformation. Due to the lack of image details, the 
	background pixels may 
	need the large displacement to find their corresponding position. 
	Since the Soft flow values are normalized for displaying, it makes the color of the face area lighter (in the 4th row). But in fact, their magnitudes are larger than other hard flows, which can be seen from the 8th row.\\
	
	\noindent\textbf{GK flow}\\
	The KG flow (also in the quarter resolution) in SCDM are displayed on the 5th and 9th rows. 
	Their magnitudes 
	are small, but they have more obvious direction information. \emph{E.g.}, 
	when the face turns left and right, it is shown in blue and red, respectively. It demonstrates that the view difference information of $C_{diff}$ has been effectively applied. 
	
	\subsection{Visualization for two components of the final generated images}
\begin{figure*} 
		\begin{center} 
			\includegraphics[width=0.95\textwidth]{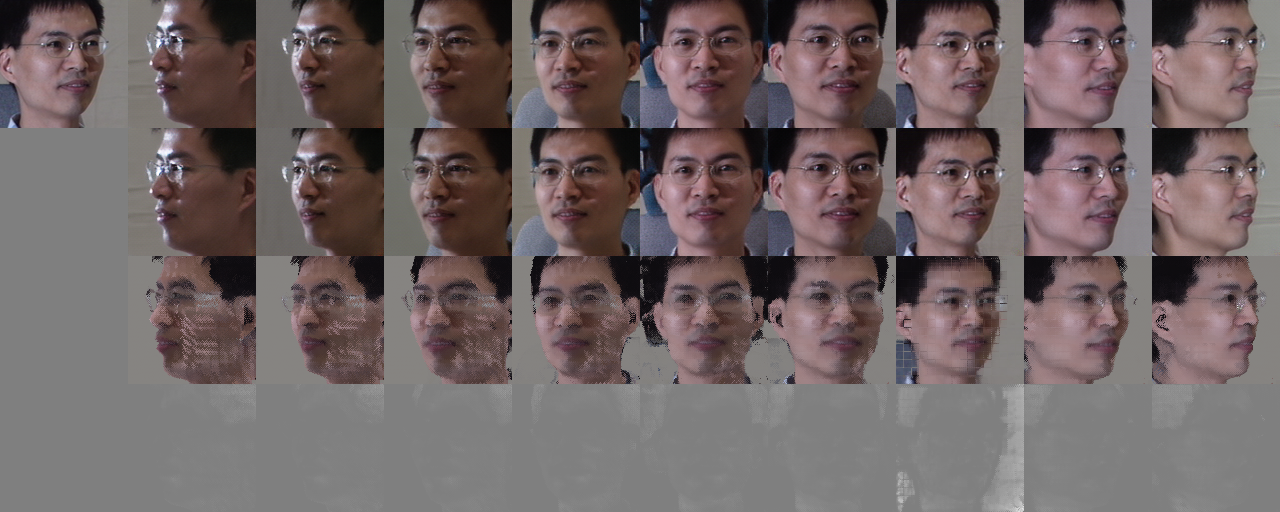}
		    \includegraphics[width=0.95\textwidth]{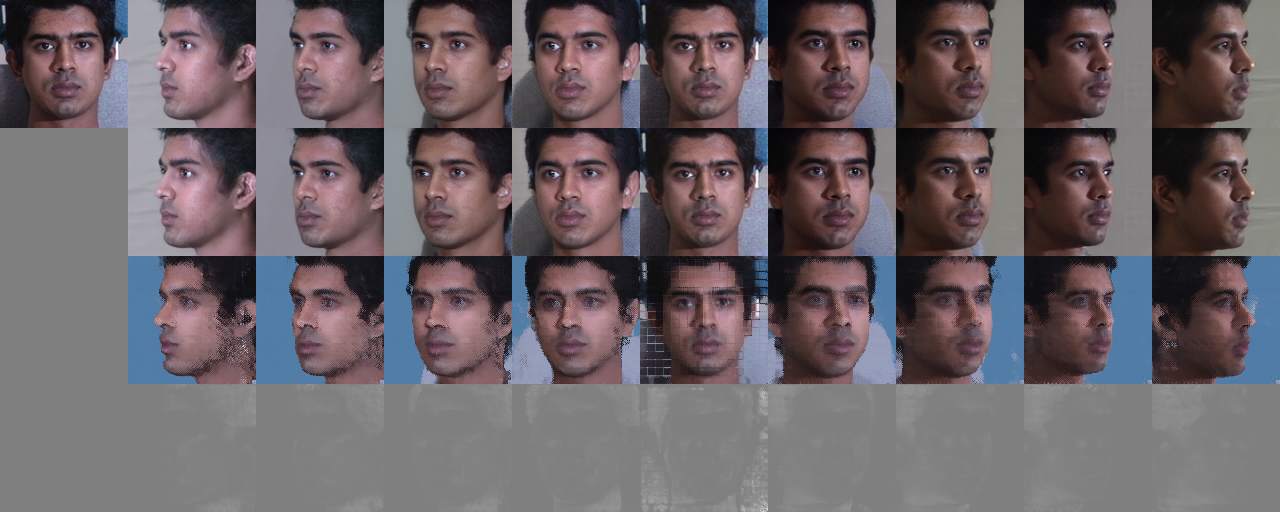}
			\includegraphics[width=0.95\textwidth]{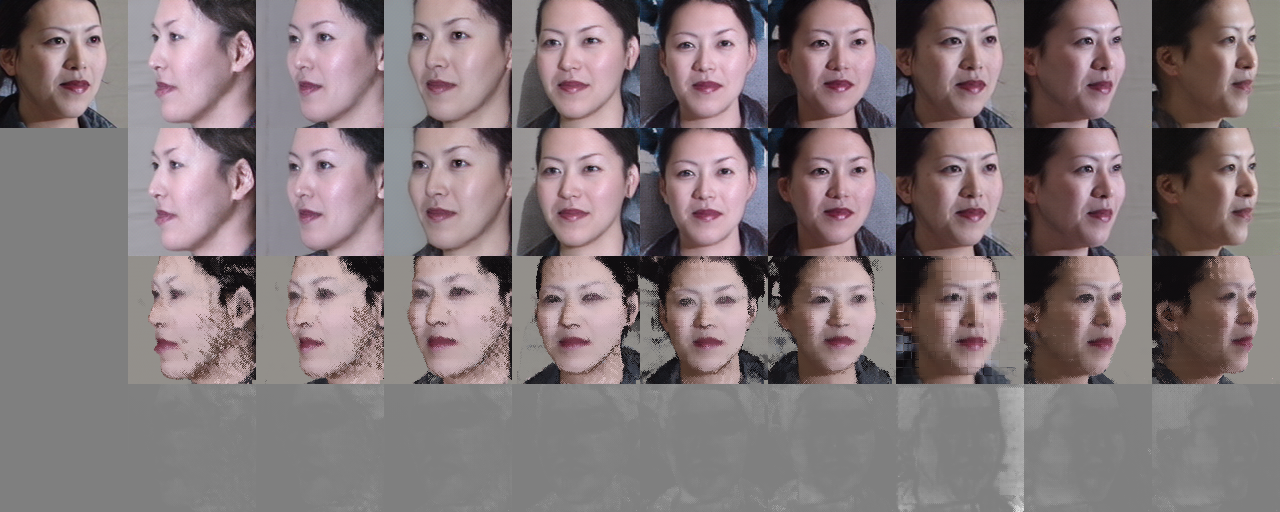}
		\end{center}
		\caption{The 1st,2dn and 3rd rows are the final generated image, and its two components of $X^g$
		and $X^{warp}$.
		The 4th row is the $mask$ for weighting. The darker the color of the $mask$, the lower the weight of $X^{warp}$.}
		\label{fig:warp}
	\end{figure*}

	Here We show 
	the final generated image, and its two components of $X^g$ and $X^{warp}$. They are combined by $1-mask$ and $mask$, respectively. Note that $X^{warp}$ clearly indicates the effectiveness of the flow, since it directly deforms the raw pixels. As shown in Figure \ref{fig:warp}, the 1st, 2nd, 3rd, and 4th rows are $X$, $X^g$, $X^{warp}$ and $mask$, respectively. 
	
	The darker the color of the $mask$, the lower the weight of $X^{warp}$
	(Gray means the value is close to 0). It can be found that $X^g$ is good enough compared to $X^{warp}$, and the weight (1-$mask$) on $X^g$ is also larger.
	
	It is observed that $X^{warp}$ maintains the brightness, color and identity of the original image to a large extent. At the same time, for the invisible areas in the source image, some areas will be missing in the deformed image $X^{warp}$. 
	Although not in high quality, 
	its view is still correct, 
	therefore, it can better assist the generation of $X^g$ in the way of DFNM.
	
	\clearpage
	
	{\small
		\bibliographystyle{ieee_fullname}
		\bibliography{egbib}

\begin{thebibliography}{10}\itemsep=-1pt

\bibitem{aubry2014seeing}
Mathieu Aubry, Daniel Maturana, Alexei~A Efros, Bryan~C Russell, and Josef
  Sivic.
\newblock Seeing 3d chairs: exemplar part-based 2d-3d alignment using a large
  dataset of cad models.
\newblock In {\em Proceedings of the IEEE conference on computer vision and
  pattern recognition}, pages 3762--3769, 2014.

\bibitem{avidan1997novel}
Shai Avidan and Amnon Shashua.
\newblock Novel view synthesis in tensor space.
\newblock In {\em Proceedings of IEEE Computer Society Conference on Computer
  Vision and Pattern Recognition}, pages 1034--1040. IEEE, 1997.

\bibitem{baker2004lucas}
Simon Baker and Iain Matthews.
\newblock Lucas-kanade 20 years on: A unifying framework.
\newblock {\em International journal of computer vision}, 56(3):221--255, 2004.

\bibitem{bao2017cvae}
Jianmin Bao, Dong Chen, Fang Wen, Houqiang Li, and Gang Hua.
\newblock Cvae-gan: fine-grained image generation through asymmetric training.
\newblock In {\em Proceedings of the IEEE international conference on computer
  vision}, pages 2745--2754, 2017.

\bibitem{brock2018large}
Andrew Brock, Jeff Donahue, and Karen Simonyan.
\newblock Large scale gan training for high fidelity natural image synthesis.
\newblock {\em arXiv preprint arXiv:1809.11096}, 2018.

\bibitem{choi2018stargan}
Yunjey Choi, Minje Choi, Munyoung Kim, Jung-Woo Ha, Sunghun Kim, and Jaegul
  Choo.
\newblock Stargan: Unified generative adversarial networks for multi-domain
  image-to-image translation.
\newblock In {\em Proceedings of the IEEE conference on computer vision and
  pattern recognition}, pages 8789--8797, 2018.

\bibitem{dosovitskiy2015learning}
Alexey Dosovitskiy, Jost Tobias~Springenberg, and Thomas Brox.
\newblock Learning to generate chairs with convolutional neural networks.
\newblock In {\em Proceedings of the IEEE conference on computer vision and
  pattern recognition}, pages 1538--1546, 2015.

\bibitem{dumoulin2016learned}
Vincent Dumoulin, Jonathon Shlens, and Manjunath Kudlur.
\newblock A learned representation for artistic style.
\newblock {\em arXiv preprint arXiv:1610.07629}, 2016.

\bibitem{esser2018variational}
Patrick Esser, Ekaterina Sutter, and Bj{\"o}rn Ommer.
\newblock A variational u-net for conditional appearance and shape generation.
\newblock In {\em Proceedings of the IEEE Conference on Computer Vision and
  Pattern Recognition}, pages 8857--8866, 2018.

\bibitem{goodfellow2014generative}
Ian Goodfellow, Jean Pouget-Abadie, Mehdi Mirza, Bing Xu, David Warde-Farley,
  Sherjil Ozair, Aaron Courville, and Yoshua Bengio.
\newblock Generative adversarial nets.
\newblock In {\em Advances in neural information processing systems}, pages
  2672--2680, 2014.

\bibitem{gross2010multi}
Ralph Gross, Iain Matthews, Jeffrey Cohn, Takeo Kanade, and Simon Baker.
\newblock Multi-pie.
\newblock {\em Image and Vision Computing}, 28(5):807--813, 2010.

\bibitem{gulrajani2017improved}
Ishaan Gulrajani, Faruk Ahmed, Martin Arjovsky, Vincent Dumoulin, and Aaron~C
  Courville.
\newblock Improved training of wasserstein gans.
\newblock In {\em Advances in neural information processing systems}, pages
  5767--5777, 2017.

\bibitem{heusel2017gans}
Martin Heusel, Hubert Ramsauer, Thomas Unterthiner, Bernhard Nessler, and Sepp
  Hochreiter.
\newblock Gans trained by a two time-scale update rule converge to a local nash
  equilibrium.
\newblock In {\em Advances in neural information processing systems}, pages
  6626--6637, 2017.

\bibitem{higgins2016beta}
Irina Higgins, Loic Matthey, Arka Pal, Christopher Burgess, Xavier Glorot,
  Matthew Botvinick, Shakir Mohamed, and Alexander Lerchner.
\newblock beta-vae: Learning basic visual concepts with a constrained
  variational framework.
\newblock 2016.

\bibitem{huang2017arbitrary}
Xun Huang and Serge Belongie.
\newblock Arbitrary style transfer in real-time with adaptive instance
  normalization.
\newblock In {\em Proceedings of the IEEE International Conference on Computer
  Vision}, pages 1501--1510, 2017.

\bibitem{isola2017image}
Phillip Isola, Jun-Yan Zhu, Tinghui Zhou, and Alexei~A Efros.
\newblock Image-to-image translation with conditional adversarial networks.
\newblock In {\em Proceedings of the IEEE conference on computer vision and
  pattern recognition}, pages 1125--1134, 2017.

\bibitem{karras2019style}
Tero Karras, Samuli Laine, and Timo Aila.
\newblock A style-based generator architecture for generative adversarial
  networks.
\newblock In {\em Proceedings of the IEEE conference on computer vision and
  pattern recognition}, pages 4401--4410, 2019.

\bibitem{kholgade20143d}
Natasha Kholgade, Tomas Simon, Alexei Efros, and Yaser Sheikh.
\newblock 3d object manipulation in a single photograph using stock 3d models.
\newblock {\em ACM Transactions on Graphics (TOG)}, 33(4):1--12, 2014.

\bibitem{kingma2014adam}
Diederik~P Kingma and Jimmy Ba.
\newblock Adam: A method for stochastic optimization.
\newblock {\em arXiv preprint arXiv:1412.6980}, 2014.

\bibitem{kingma2013auto}
Diederik~P Kingma and Max Welling.
\newblock Auto-encoding variational bayes.
\newblock {\em arXiv preprint arXiv:1312.6114}, 2013.

\bibitem{li2019positional}
Boyi Li, Felix Wu, Kilian~Q Weinberger, and Serge Belongie.
\newblock Positional normalization.
\newblock In {\em Advances in Neural Information Processing Systems}, pages
  1622--1634, 2019.

\bibitem{lim2017geometric}
Jae~Hyun Lim and Jong~Chul Ye.
\newblock Geometric gan.
\newblock {\em arXiv preprint arXiv:1705.02894}, 2017.

\bibitem{lucas1981iterative}
Bruce~D Lucas, Takeo Kanade, et~al.
\newblock An iterative image registration technique with an application to
  stereo vision.
\newblock 1981.

\bibitem{miyato2018spectral}
Takeru Miyato, Toshiki Kataoka, Masanori Koyama, and Yuichi Yoshida.
\newblock Spectral normalization for generative adversarial networks.
\newblock {\em arXiv preprint arXiv:1802.05957}, 2018.

\bibitem{nam2018batch}
Hyeonseob Nam and Hyo-Eun Kim.
\newblock Batch-instance normalization for adaptively style-invariant neural
  networks.
\newblock In {\em Advances in Neural Information Processing Systems}, pages
  2558--2567, 2018.

\bibitem{nguyen2019hologan}
Thu Nguyen-Phuoc, Chuan Li, Lucas Theis, Christian Richardt, and Yong-Liang
  Yang.
\newblock Hologan: Unsupervised learning of 3d representations from natural
  images.
\newblock In {\em Proceedings of the IEEE International Conference on Computer
  Vision}, pages 7588--7597, 2019.

\bibitem{odena2017conditional}
Augustus Odena, Christopher Olah, and Jonathon Shlens.
\newblock Conditional image synthesis with auxiliary classifier gans.
\newblock In {\em International conference on machine learning}, pages
  2642--2651, 2017.

\bibitem{park2017transformation}
Eunbyung Park, Jimei Yang, Ersin Yumer, Duygu Ceylan, and Alexander~C Berg.
\newblock Transformation-grounded image generation network for novel 3d view
  synthesis.
\newblock In {\em Proceedings of the ieee conference on computer vision and
  pattern recognition}, pages 3500--3509, 2017.

\bibitem{park2019semantic}
Taesung Park, Ming-Yu Liu, Ting-Chun Wang, and Jun-Yan Zhu.
\newblock Semantic image synthesis with spatially-adaptive normalization.
\newblock In {\em Proceedings of the IEEE Conference on Computer Vision and
  Pattern Recognition}, pages 2337--2346, 2019.

\bibitem{parkhi2015deep}
Omkar~M Parkhi, Andrea Vedaldi, and Andrew Zisserman.
\newblock Deep face recognition.
\newblock 2015.

\bibitem{rematas2016novel}
Konstantinos Rematas, Chuong~H Nguyen, Tobias Ritschel, Mario Fritz, and Tinne
  Tuytelaars.
\newblock Novel views of objects from a single image.
\newblock {\em IEEE transactions on pattern analysis and machine intelligence},
  39(8):1576--1590, 2016.

\bibitem{ronneberger2015u}
Olaf Ronneberger, Philipp Fischer, and Thomas Brox.
\newblock U-net: Convolutional networks for biomedical image segmentation.
\newblock In {\em International Conference on Medical image computing and
  computer-assisted intervention}, pages 234--241. Springer, 2015.

\bibitem{siarohin2018deformable}
Aliaksandr Siarohin, Enver Sangineto, St{\'e}phane Lathuiliere, and Nicu Sebe.
\newblock Deformable gans for pose-based human image generation.
\newblock In {\em Proceedings of the IEEE Conference on Computer Vision and
  Pattern Recognition}, pages 3408--3416, 2018.

\bibitem{simonyan2014very}
Karen Simonyan and Andrew Zisserman.
\newblock Very deep convolutional networks for large-scale image recognition.
\newblock {\em arXiv preprint arXiv:1409.1556}, 2014.

\bibitem{sohn2015learning}
Kihyuk Sohn, Honglak Lee, and Xinchen Yan.
\newblock Learning structured output representation using deep conditional
  generative models.
\newblock In {\em Advances in neural information processing systems}, pages
  3483--3491, 2015.

\bibitem{sun2018multi}
Shao-Hua Sun, Minyoung Huh, Yuan-Hong Liao, Ning Zhang, and Joseph~J Lim.
\newblock Multi-view to novel view: Synthesizing novel views with self-learned
  confidence.
\newblock In {\em Proceedings of the European Conference on Computer Vision
  (ECCV)}, pages 155--171, 2018.

\bibitem{tian2018cr}
Yu Tian, Xi Peng, Long Zhao, Shaoting Zhang, and Dimitris~N Metaxas.
\newblock Cr-gan: learning complete representations for multi-view generation.
\newblock {\em arXiv preprint arXiv:1806.11191}, 2018.

\bibitem{tran2017disentangled}
Luan Tran, Xi Yin, and Xiaoming Liu.
\newblock Disentangled representation learning gan for pose-invariant face
  recognition.
\newblock In {\em Proceedings of the IEEE conference on computer vision and
  pattern recognition}, pages 1415--1424, 2017.

\bibitem{ulyanov2016instance}
Dmitry Ulyanov, Andrea Vedaldi, and Victor Lempitsky.
\newblock Instance normalization: The missing ingredient for fast stylization.
\newblock {\em arXiv preprint arXiv:1607.08022}, 2016.

\bibitem{vaswani2017attention}
Ashish Vaswani, Noam Shazeer, Niki Parmar, Jakob Uszkoreit, Llion Jones,
  Aidan~N Gomez, {\L}ukasz Kaiser, and Illia Polosukhin.
\newblock Attention is all you need.
\newblock In {\em Advances in neural information processing systems}, pages
  5998--6008, 2017.

\bibitem{wang2018high}
Ting-Chun Wang, Ming-Yu Liu, Jun-Yan Zhu, Andrew Tao, Jan Kautz, and Bryan
  Catanzaro.
\newblock High-resolution image synthesis and semantic manipulation with
  conditional gans.
\newblock In {\em Proceedings of the IEEE conference on computer vision and
  pattern recognition}, pages 8798--8807, 2018.

\bibitem{wang2018non}
Xiaolong Wang, Ross Girshick, Abhinav Gupta, and Kaiming He.
\newblock Non-local neural networks.
\newblock In {\em Proceedings of the IEEE conference on computer vision and
  pattern recognition}, pages 7794--7803, 2018.

\bibitem{wang2004image}
Zhou Wang, Alan~C Bovik, Hamid~R Sheikh, and Eero~P Simoncelli.
\newblock Image quality assessment: from error visibility to structural
  similarity.
\newblock {\em IEEE transactions on image processing}, 13(4):600--612, 2004.

\bibitem{xiao2018elegant}
Taihong Xiao, Jiapeng Hong, and Jinwen Ma.
\newblock Elegant: Exchanging latent encodings with gan for transferring
  multiple face attributes.
\newblock In {\em Proceedings of the European conference on computer vision
  (ECCV)}, pages 168--184, 2018.

\bibitem{xu2019view}
Xiaogang Xu, Ying-Cong Chen, and Jiaya Jia.
\newblock View independent generative adversarial network for novel view
  synthesis.
\newblock In {\em Proceedings of the IEEE International Conference on Computer
  Vision}, pages 7791--7800, 2019.

\bibitem{yin2020novel}
Mingyu Yin, Li Sun, and Qingli Li.
\newblock Novel view synthesis on unpaired data by conditional deformable
  variational auto-encoder.
\newblock {\em arXiv preprint arXiv:2007.10618}, 2020.

\bibitem{zhang2018unreasonable}
Richard Zhang, Phillip Isola, Alexei~A Efros, Eli Shechtman, and Oliver Wang.
\newblock The unreasonable effectiveness of deep features as a perceptual
  metric.
\newblock In {\em Proceedings of the IEEE conference on computer vision and
  pattern recognition}, pages 586--595, 2018.

\bibitem{zheng2019disentangling}
Zhilin Zheng and Li Sun.
\newblock Disentangling latent space for vae by label relevant/irrelevant
  dimensions.
\newblock In {\em Proceedings of the IEEE Conference on Computer Vision and
  Pattern Recognition}, pages 12192--12201, 2019.

\bibitem{zhou2016view}
Tinghui Zhou, Shubham Tulsiani, Weilun Sun, Jitendra Malik, and Alexei~A Efros.
\newblock View synthesis by appearance flow.
\newblock In {\em European conference on computer vision}, pages 286--301.
  Springer, 2016.

\bibitem{zhu2017unpaired}
Jun-Yan Zhu, Taesung Park, Phillip Isola, and Alexei~A Efros.
\newblock Unpaired image-to-image translation using cycle-consistent
  adversarial networks.
\newblock In {\em Proceedings of the IEEE international conference on computer
  vision}, pages 2223--2232, 2017.

\bibitem{zhu2017toward}
Jun-Yan Zhu, Richard Zhang, Deepak Pathak, Trevor Darrell, Alexei~A Efros,
  Oliver Wang, and Eli Shechtman.
\newblock Toward multimodal image-to-image translation.
\newblock In {\em Advances in neural information processing systems}, pages
  465--476, 2017.

\end{thebibliography}
	}

\end{document}